\documentclass{article}

\usepackage[preprint]{neurips_2019}

\usepackage[utf8]{inputenc} % allow utf-8 input
\usepackage[T1]{fontenc}    % use 8-bit T1 fonts
\usepackage{hyperref}       % hyperlinks
\usepackage{url}            % simple URL typesetting
\usepackage{booktabs}       % professional-quality tables
\usepackage{amsfonts}       % blackboard math symbols
\usepackage{nicefrac}       % compact symbols for 1/2, etc.
\usepackage{microtype}      % microtypography
\usepackage{graphicx}
\usepackage{amsmath}
\usepackage{xcolor}
\usepackage{caption}
\usepackage{subcaption} 

\usepackage{tabularx}
\usepackage{nccmath}

\usepackage[utf8]{inputenc}
\usepackage[english]{babel}
\usepackage{amsthm}
\usepackage{cleveref}

\newtheorem{theorem}{Theorem}[section]
\newtheorem*{theorem*}{Theorem}
\newtheorem{corollary}{Corollary}[theorem]
\newtheorem{lemma}[theorem]{Lemma}

\DeclareMathOperator*{\E}{\mathbb{E}}
\DeclareMathOperator*{\argmax}{arg\,max}
\DeclareMathOperator*{\argmin}{arg\,min}
\DeclareMathOperator*{\LSE}{\text{L}{\Sigma}\text{E}}
\DeclareMathOperator*{\softmax}{\text{$\boldsymbol\sigma$}}  % {\sigma}

\title{Importance Weighted Hierarchical \\ Variational Inference}

\author{%
  Artem Sobolev \\
  Samsung AI Center Moscow, Russia \\
  \texttt{asobolev@bayesgroup.ru} \\
  \And
  Dmitry Vetrov \\
  NRU HSE\thanks{Samsung-HSE Joint Lab, National Research University Higher School of Economics}, Moscow, Russia \\
  Samsung AI Center Moscow, Russia \\
}

\begin{document}
\maketitle
\begin{abstract}
Variational Inference is a powerful tool in the Bayesian modeling toolkit, however, its effectiveness is determined by the expressivity of the utilized variational distributions in terms of their ability to match the true posterior distribution. In turn, the expressivity of the variational family is largely limited by the requirement of having a tractable density function.
To overcome this roadblock, we introduce a new family of variational upper bounds on a marginal log density in the case of hierarchical models (also known as latent variable models). We then give an upper bound on the Kullback-Leibler divergence and derive a family of increasingly tighter variational lower bounds on the otherwise intractable standard evidence lower bound for hierarchical variational distributions, enabling the use of more expressive approximate posteriors. We show that previously known methods, such as Hierarchical Variational Models, Semi-Implicit Variational Inference and Doubly Semi-Implicit Variational Inference can be seen as special cases of the proposed approach, and empirically demonstrate superior performance of the proposed method in a set of experiments.
\end{abstract}

\section{Introduction}
Bayesian Inference is an important statistical tool. However, exact inference is possible only in a small class of conjugate problems, and for many practically interesting cases, one has to resort to Approximate Inference techniques. Variational Inference \citep{Hinton:1993:KNN:168304.168306,waterhouse1996bayesian,wainwright2008graphical} being one of them is an efficient and scalable approach that gained a lot of interest in recent years due to advances in Neural Networks.

However, the efficiency and accuracy of Variational Inference heavily depend on how close an approximate posterior is to the true posterior. As a result, Neural Networks' universal approximation abilities and great empirical success propelled a lot of interest in employing them as powerful sample generators \citep{NIPS2016_6066,goodfellow2014generative,MACKAY199573} that are trained to output samples from approximate posterior when fed some standard noise as input.
% \citet{kingma2013auto} used neural networks to amortize inference; however, this alone did not increase approximate posterior's expressivity.
Unfortunately, a significant obstacle on this direction is a need for a tractable density $q(z \mid x)$, which in general requires intractable integration. A theoretically sound approach then is to give tight lower bounds on the intractable term -- the differential entropy of $q(z|x)$, which is easy to recover from upper bounds on the marginal log-density. One such bound was introduced by \citet{agakov2004auxiliary}; however it's tightness depends on the auxiliary variational distribution. \citet{yin2018semi} suggested a multisample loss, whose tightness is controlled by the number of samples.

In this paper we consider hierarchical variational models \citep{ranganath2016hierarchical,salimans2015markov,agakov2004auxiliary} where the approximate posterior $q(z \mid x)$ is represented as a mixture of tractable distributions $q(z | \psi, x)$ over arbitrarily complicated mixing distribution $q(\psi | x)$: ${q(z | x)} = \int {q(z | \psi, x)}{ q(\psi | x)} d\psi$. We show that such variational models contain \emph{semi-implicit models}, first studied by \citet{yin2018semi}. To overcome the need for the closed-form marginal density $q(z | x)$ we then propose a novel family of tighter bounds on the marginal log-likelihood $\log p(x)$, which can be shown to generalize many previously known bounds: Hierarchical Variational Models \citep{ranganath2016hierarchical} also known as auxiliary VAE bound \citep{pmlr-v48-maaloe16}, Semi-Implicit Variational Inference \citep{yin2018semi} and Doubly Semi-Implicit Variational Inference \citep{molchanov2018doubly}. At the core of our work lies a novel variational upper bound on the marginal log-likelihood, which we combine with previously known lower bound \citep{burda2015importance} to give novel upper bound on the Kullback-Leibler (KL) divergence between hierarchical models, and apply it to the evidence lower bound (ELBO) to enable hierarchical approximate posteriors and/or priors. Finally, our method can be combined with the multisample bound of \citet{burda2015importance} to tighten the marginal log-likelihood estimate further.

\section{Background} \label{sec:bkg}
Having a hierarchical model $p_\theta(x) = \int p_\theta(x \mid z) p_\theta(z) dz$ for observable objects $x$, we are interested in two tasks: inference and learning. The problem of Bayesian inference is that of finding the true posterior distribution $p_\theta(z \mid x)$, which is often intractable and thus is approximated by some $q_\phi(z \mid x)$. The problem of learning is that of finding parameters $\theta$ s.t. the marginal model distribution $p_\theta(x)$ approximates the true data-generating process of $x$ as good as possible, typically in terms of KL-divergence, which leads to the Maximum Likelihood Estimation problem.

Variational Inference provides a way to solve both tasks simultaneously by lower-bounding the intractable marginal log-likelihood $\log p_\theta(x)$ with the Evidence Lower Bound (ELBO) using the posterior approximation $q_\phi(z \mid x)$:
$$
\log p_\theta(x) \ge \E_{q_\phi(z \mid x)} \log \frac{p_\theta(x, z)}{q_\phi(z \mid x)}.
$$
The bound requires analytically tractable densities for both $q_\phi(z \mid x)$ and $p_\theta(z)$. The gap between the marginal log-likelihood and the bound is equal to $D_{KL}(q_\phi(z \mid x) \mid\mid p_\theta(z \mid x))$, which acts as a \textit{regularization} preventing the true posterior $p(z|x)$ from deviating too far away from the approximate one $q(z|x)$, thus limiting the expressivity of the marginal distribution $p(x)$. \citet{burda2015importance} proposed a family of tighter multisample bounds, generalizing the ELBO. We call it the IWAE bound:
$$
\log p_\theta(x) \ge \E_{q_\phi(z_{1:M} \mid x)} \log \frac{1}{M} \sum_{m=1}^M \frac{p_\theta(x, z_m)}{q_\phi(z_m \mid x)}.
$$
Where from now on we write $q_\phi(z_{1:M} | x) = \prod_{m=1}^{M} q_\phi(z_m | x)$ for brevity. This bound has been shown \citep{domke2018importance} to be a tractable lower bound on ELBO for a variational distribution that has been obtained by a certain scoring procedure. However, the price of this increased tightness is higher computation complexity, especially in terms of the number of evaluations of the joint distribution $p_\theta(x, z)$, and thus we might want to come up with a more expressive posterior approximation to be used in the ELBO -- a special case of $M=1$.

In the direction of improving the single-sample ELBO it was proposed \citep{agakov2004auxiliary,salimans2015markov,pmlr-v48-maaloe16,ranganath2016hierarchical} to use a hierarchical variational model (HVM) for $q(z \mid x) = \int {q(z \mid x, \psi)} {q(\psi \mid x)} d\psi$ with explicit joint density ${q(z, \psi \mid x)}$, where $\psi$ is auxiliary random variables. To overcome the intractability of the density ${q(z \mid x)}$, a variational lower bound on the ELBO is proposed. The tightness of the bound is controlled by the auxiliary variational distribution ${\tau(\psi \mid x, z)}$:
\begin{align} \label{eq:hvm-elbo}
\log p_\theta(x) \ge \E_{q_\phi(z, \psi \mid x)} \left[ \log p_\theta(x, z) - \log \frac{q_\phi(z, \psi \mid x)}{\tau(\psi \mid x, z)} \right]
\end{align}

Recently \citet{yin2018semi} introduced \emph{semi-implicit models}: hierarchical models $q(z \mid x) = \int q_\phi(z \mid \psi, x) q_\phi(\psi \mid x) d\psi$ with implicit but reparametrizable ${q_\phi(\psi \mid x)}$ and explicit ${q_\phi(z \mid \psi, x)}$, and suggested the following surrogate objective, which was later shown to be a lower bound (the SIVI bound) for all finite $K$ by \citet{molchanov2018doubly}:
\begin{align} \label{eq:sivi-elbo}
\log p_\theta(x)
\ge
\E_{q_\phi(z, \psi_0 \mid x)}
\E_{q_\phi(\psi_{1:K} \mid x)}
\log \frac{p_\theta(x, z)}{\tfrac{1}{K+1} \sum_{k=0}^K q_\phi(z | \psi_k, x)}
\end{align}
SIVI also can be generalized into a multisample bound similarly to the IWAE bound \citep{burda2015importance} in an efficient way by reusing samples $\psi_{1:K}$ for different $z_m$:
\begin{align} \label{eq:sivi-reused}
\log p(x)
\ge
\E
%\left[
\log \tfrac{1}{M} \sum_{m=1}^M \frac{p_\theta(x, z_{m})}{\tfrac{1}{K + 1} \sum_{k=0}^{K} q_\phi(z_{m} | x, \psi_{m,k})}
%\right]
\end{align}
Where the expectation is taken over ${q(\psi_{1:M,0}, z_{1:M} \mid x)}$ and $\psi_{m,1:K} = \psi_{1:K} \sim q(\psi_{1:K} \mid x)$ is the same set of $K$ i.i.d. random variables for all $m$\footnote{One could also include all $\psi_{1:M,0}$ into the set of reused samples $\psi$, expanding its size to $M+K$.}. Importantly, this estimator has $O(M+K)$ sampling complexity for $\psi$, unlike the naive approach, leading to $O(M K+M)$ sampling complexity. We will get back to this discussion in \cref{sec:multisample-and-reuse}.

\subsection{SIVI Insights} \label{sec:sivi-insights}
Here we outline SIVI's points of weaknesses and identify certain traits that make it possible to generalize the method and bridge it with the prior work.

First, note that both SIVI bounds \eqref{eq:sivi-elbo} and \eqref{eq:sivi-reused} use samples from ${q_\phi(\psi_{1:K} \mid x)}$ to describe $z$, and in high dimensions one might expect that such "uninformed" samples would miss most of the time, resulting in near-zero likelihood ${q(z \mid \psi_k, x)}$ and thus reducing the effective sample size. Therefore it is expected that in higher dimensions it would take many samples to accurately cover the regions high probability of $q(\psi \mid z, x)$ for the given $z$. Instead, ideally, we would like to target such regions directly while keeping the lower bound guarantees.

Another important observation that we'll make use of is that many such semi-implicit models can be equivalently reformulated as a mixture of two explicit distributions: due to reparametrizability of $q_\phi(\psi \mid x)$ we have $\psi = g_\phi(\varepsilon \mid x)$ for some $\varepsilon \sim q(\varepsilon)$ with tractable density. We can then consider an equivalent hierarchical model $q_\phi(z) = {\int q_\phi(z \mid g_\phi(\varepsilon \mid x), x) q(\varepsilon) d\varepsilon}$ that first samples $\varepsilon$ from some simple distribution, transforms this sample $\varepsilon$ into $\psi$ and then generates samples from $q_\phi(z \mid \psi, x)$. Thus from now on we'll assume both $q_\phi(z \mid \psi, x)$ and $q_\phi(\psi \mid x)$ have tractable density, yet $q_\phi(z \mid \psi, x)$ can depend on $\psi$ in an arbitrarily complex way, making analytic marginalization intractable. % However, one might be interested in hierarchical models with implicit non-reparametrizable mixing distribution $q(\psi|x)$ such as "aggregates posteriors" \citep{molchanov2018doubly}.

% In the next section we derive a general variational upper bound on the marginal log density $\log q_\phi(z \mid x)$, and utilize this bound to come up with a more general lower bound on marginal log-likelihood $\log p_\theta(x)$. We then show that this bound generalizes and bridges many previous works like HVM and SIVI.

\section{Importance Weighted Hierarchical Variational Inference}
Having intractable $\log q_\phi(z \mid x)$ as a source of our problems, we seek a tractable and efficient \textit{upper} bound, which is provided by the following theorem:

\begin{theorem*}[Marginal log density upper bound]
For any $q(z, \psi \mid x)$, $K \in \mathbb{N}_0$ and $\tau(\psi \mid z, x)$ (under some regularity conditions) consider the following
$$
\mathcal{U}_{K} = \E_{q(\psi_0 \mid x, z)} \E_{\tau(\psi_{1:K} \mid z, x)} \log \left( \frac{1}{K+1} \sum_{k=0}^{K} \frac{q(z, \psi_k \mid x)}{\tau(\psi_k \mid z, x)} \right)
$$
Then the following holds:
\begin{enumerate}
    \item $\mathcal{U}_{K} \ge \log q(z \mid x)$
    \item $\mathcal{U}_{K} \ge \mathcal{U}_{K + 1}$
    \item $\lim\limits_{K \to \infty} \mathcal{U}_K = \log q(z \mid x)$
\end{enumerate}
\begin{proof}
See Appendix for \Cref{thm:upperboundthm}.
\end{proof}
\end{theorem*}

The proposed upper bound provides a variational alternative to MCMC-based upper bounds \citep{Grosse2015SandwichingTM} and complements the standard Importance Weighted stochastic lower bound of \citet{burda2015importance} on the marginal log density:
$$
\mathcal{L}_{K}
=
\E_{\tau(\psi_{1:K} \mid z, x)} \log \left( \frac{1}{K} \sum_{k=1}^{K} \frac{q(z, \psi_k \mid x)}{\tau(\psi_k \mid z, x)} \right)
\le
\log q(z \mid x)
$$

\subsection{Upper Bound on KL divergence between hierarchical models}
We now apply these bounds to marginal log-densities, appearing in KL divergence in case of both ${q(z | x)} = \int {q(z, \psi | x)} d\psi$ and $p(z) = \int {p(z, \zeta)} d\zeta$ being different (potentially structurally) hierarchical models. This results in a novel upper bound on KL divergence with auxiliary variational distributions $\tau(\psi \mid x, z)$ and $\rho(\zeta \mid z)$:
\begin{align} 
D_{KL}(q(z \mid x) \mid\mid \; p(z))
&\le
\E_{q(z, \psi_0 \mid x)}
\E_{\tau(\psi_{1:K} \mid x, z)}
\E_{\rho(\zeta_{1:L} \mid z)} 
\Biggl[
\log
\frac{
    \tfrac{1}{K + 1} \sum_{k=0}^{K} \tfrac{q(z, \psi_k \mid x)}{\tau(\psi_k \mid x, z)}
}{
    \tfrac{1}{L} \sum_{l=1}^{L} \frac{p(z, \zeta_l)}{\rho(\zeta_l \mid z)}
}
\Biggr]
\label{eq:kl-upper-bound}
\end{align}
Crucially, in \eqref{eq:kl-upper-bound} we merged expectations over $q_\phi(z | x)$ and $q_\phi(\psi_0 | x, z)$ into one expectation over the joint distribution $q_\phi(\psi_0, z | x)$, which admits a more favorable factorization into ${q_\phi(\psi_0 | x) q_\phi(z | x, \psi_0)}$, and samples from the later are easy to simulate for the Monte Carlo-based estimation.

One can also give lower bounds in the same setting at \eqref{eq:kl-upper-bound}. However, the main focus of this paper is on lower bounding the marginal log-likelihood, so we leave this discussion to \cref{sec:kl-lower-bound}.

\subsection{Tractable lower bounds on marginal log-likelihood with hierarchical proposal}
The proposed upper bound \eqref{eq:kl-upper-bound} allows us to lower bound the otherwise intractable ELBO in case of hierarchical $q(z \mid x)$ and $p(z)$, leading to \textbf{Importance Weighted Hierarchical Variational Inference} (IWHVI) lower bound:
\begin{align} \label{eq:bound}
\log p(x)
& \ge
\E_{q(z \mid x)} \log \frac{p(x, z)}{q(z \mid x)}
%=
%\E_{q(z \mid x)} \left[ \log p(x \mid z) + \log p(z) - \log q(z \mid x) \right]
\ge
\E_{q(z, \psi_0 \mid x)}
\E_{\tau(\psi_{1:K} \mid z, x)}
\E_{\rho(\zeta_{1:L} \mid z)}
\log
\frac{p(x \mid z) \frac{1}{L} \sum_{k=1}^{L} \frac{p(z, \zeta_l)}{\rho(\zeta_l \mid z)} }{ \frac{1}{K + 1} \sum_{k=0}^{K} \frac{q(z, \psi_k \mid x)}{\tau(\psi_k \mid z, x)} }
\end{align}
% Where the expectation is taken over ${q(z, \psi_0 \mid x)} {\tau(\psi_{1:K} \mid z)} {\rho(\zeta_{1:K} \mid z)}$.
This bound introduces two additional auxiliary variational distributions $\rho$ and $\tau$ that are learned by maximizing the bound w.r.t. their parameters, tightening the bound. While the optimal distributions are\footnote{This choice makes bounds $\mathcal{U}_K$ and $\mathcal{L}_K$ equal to the marginal log-density.} $\tau(\psi \mid z, x) = q(\psi \mid z, x)$ and $\rho(\zeta \mid z) = p(\zeta \mid z)$, one can see that some particular choices of these distributions and hyperparameters $K,L$ render previously known methods like DSIVI, SIVI and HVM as special cases (see \cref{sec:special-cases}).

The bound \eqref{eq:bound} can be seen as variational generalization of SIVI \eqref{eq:sivi-elbo} or multisample generalization of HVM \eqref{eq:hvm-elbo}. Therefore it has capacity to better estimate the true ELBO, reducing the gap and its regularizational effect, which should lead to more expressive variational approximations $q_\phi(z|x)$.

\section{Multisample Extensions}
Multisample bounds similar to the proposed one have already been studied extensively. In this section, we relate our results to such prior work.

\subsection{Multisample Bound and Complexity}\label{sec:multisample-and-reuse}
In this section we generalize the bound \eqref{eq:bound} further in a way similar to the IWAE multisample bound \citep{burda2015importance} (\Cref{th:iwae-bound}), leading to the \textbf{Doubly Importance Weighted Hierarchical Variational Inference} (\mbox{DIWHVI}):
\begin{align} \label{eq:iw-bound}
\log p(x)
\ge
\E
%\left[
\log \frac{1}{M} \sum_{m=1}^M \frac{p(x | z_{m}) \frac{1}{L} \sum_{l=1}^{L} \frac{p(z_{m}, \zeta_{m,l})}{\rho(\zeta_{m,l} \mid z_m)}}{\frac{1}{K + 1} \sum_{k=0}^{K} \frac{q(z_{m}, \psi_{m,k} \mid x)}{\tau(\psi_{m,k} \mid z_{m}, x)}}
%\right]
\end{align}
Where the expectation is taken over the same generative process as in \cref{eq:bound}, independently repeated $M$ times:
\begin{enumerate}
    \item Sample $\psi_{m,0} \sim q(\psi \mid x)$ for $1 \le m \le M$
    \item Sample $z_{m} \sim q(z \mid x_n, \psi_{m,0})$ for $1 \le m \le M$
    \item Sample $\psi_{m,k} \sim \tau(\psi \mid z_{m}, x)$ for $1 \le m \le M$ and $1 \le k \le K$
    \item Sample $\zeta_{m,l} \sim \rho(\zeta \mid z_{m})$ for $1 \le m \le M$ and $1 \le l \le L$
\end{enumerate}

The price of the tighter bound \eqref{eq:iw-bound} is quadratic sample complexity: it requires $M (1 + K)$ samples of $\psi$ and $M L$ samples of $\zeta$. Unfortunately, the \mbox{DIWHVI} cannot benefit from the sample reuse trick of the SIVI that leads to the bound \eqref{eq:sivi-reused}. The reason for this is that the bound \eqref{eq:iw-bound} requires all terms in the outer denominator (the $\log q_\phi(z \mid x)$ estimate) to use the same distribution $\tau$, whereas by its very nature it should be very different for different $z_m$. A viable option, though, is to consider a multisample-conditioned ${\tau(\psi \mid z_{1:M})}$ that is invariant to permutations of $z$. We leave a more detailed investigation to a future work.

Runtime-wise when compared to the multisample \mbox{SIVI} \eqref{eq:sivi-reused} the \mbox{DIWHVI} requires additional $O(M)$ passes to generate $\tau(\psi \mid x, z_m)$ distributions. However, since the SIVI requires a much larger number of samples $K$ to reach the same level of accuracy (see \cref{sec:toy-exp}) that are all then passed through a network to generate $q_\phi(z_m \mid x, \psi_{mk})$ distributions, the extra $\tau$ computation is likely to either bear a minor overhead, or be completely justified by reduced $K$. This is particularly true in the \mbox{IWHVI} case ($M=1$) where IWHVI's single extra pass that generates $\tau(\psi|x,z)$ is dominated by $K+1$ passes that generate $q(z|x, \psi_k)$.

% That said, this scheme still requires one to score each $z_m$ against each $q(z \mid \psi_k)$, leading to $O(M K)$ asymptotic computational complexity, same as for DIWHVI.

\subsection{Signal to Noise Ratio}\label{sec:snr-intro}
\citet{Rainforth2018TighterVB} have shown that multisample bounds \citep{burda2015importance,nowozin2018debiasing} behave poorly during the training phase, having more noisy inference network's gradient estimates, which manifests itself in decreasing Signal-to-Noise Ratio (SNR) as the number of samples increases. This raises a natural concern whether the same happens in the proposed model as $K$ increases. \citet{tucker2018doubly} have shown that upon a careful examination a REINFORCE-like \citep{Williams92simplestatistical} term can be seen in the gradient estimate, and REINFORCE is known for its typically high variance \citep{pmlr-v32-rezende14}. Authors further suggest to apply the reparametrization trick \mbox{\citep{kingma2013auto}} the second time to obtain a reparametrization-based gradient estimate, which is then shown to solve the decreasing SNR problem. The same reasoning can be applied to our bound, and we provide further details and experiments in \cref{sec:snr-details}, developing an IWHVI-DReG gradient estimator. We conclude that the problem of decreasing SNR exists in our bound as well, and is mitigated by the proposed gradient estimator.

\subsection{Debiasing the bound}\label{sec:jhvi-intro}
\citet{nowozin2018debiasing} has shown that the standard IWAE can be seen as a biased estimate of the marginal log-likelihood with the bias of order $O(1/M)$. They then suggested to use Generalized Jackknife of $d$-th order to reuse these $M$ samples and come up with an estimator with a smaller bias of order $O(1/M^d)$ at the cost of higher variance and losing lower bound guarantees. Again, the same idea can be applied to our estimate; we leave further details to \cref{sec:jhvi-details}. We conclude that this way one can obtain better estimates of the marginal log-density, however since there is no guarantee that the obtained estimator gives an upper or a lower bound, we chose not to use it in experiments.

\section{Related Work} \label{sec:relwork}
More expressive variational distributions have been under an active investigation for a while. While we have focused our attention to approaches employing hierarchical models via bounds, there are many other approaches, roughly falling into two broad classes.

One possible approach is to augment some standard $q(z \mid x)$ with help of copulas \citep{tran2015copula}, mixtures \citep{guo2016boosting}, or invertible transformations with tractable Jacobians also known as normalizing flows \citep{rezende2015variational,NIPS2016_6581,DBLP:journals/corr/DinhSB16,DBLP:conf/nips/PapamakariosMP17}, all while preserving the tractability of the density. \citet*{kingma2018glow} have demonstrated that flow-based models are able to approximate complex high-dimensional distributions of real images, but the requirement for invertibility might lead to inefficiency in parameters usage and does not allow for abstraction as one needs to preserve dimensions.

An alternative direction is to embrace implicit distributions that one can only sample from, and overcome the need for tractable density using bounds or estimates \citep{huszar2017variational}. Methods based on estimates \citep{mescheder2017adversarial,shi2017kernel}, for example, via the Density Ratio Estimation trick \citep{goodfellow2014generative,uehara2016generative,mohamed2016learning}, typically estimate the densities indirectly utilizing an auxiliary critic and hide dependency on variational parameters $\phi$, hence biasing the optimization procedure.
\citet{titsias2018unbiased} have shown that in the gradient-based ELBO optimization in case of a hierarchical model with tractable $q_\phi(z\mid\psi)$ and $q_\phi(\psi)$ one does not need the marginal log density $\log q_\phi(z \mid x)$ per se, only its gradient, which can be estimated using MCMC. Major disadvantage of these methods is that they either lose bound guarantees or make its evaluation intractable and thus cannot be combined with multisample bounds during the evaluation phase.

% \citet{ranganath2016hierarchical} proposed Hierarchical Variational Models (HVM) as more expressive variational approximations concurrently with the same idea of augmenting VAE with auxiliary variables \cite{agakov2004auxiliary} by \citet{pmlr-v48-maaloe16}. To cope with an intractable entropy of the marginal density $q(z \mid x)$ they introduced a variational lower bound on the entropy in hierarchical model with intractable ${q(z \mid x)} = \int {q(z \mid \psi, x)} {q(\psi \mid x)} d\psi$, giving an upper bound for the KL-divergence in a hierarchical case. \citet{yin2018semi} introduced Semi-Implicit Variational Inference (SIVI) for hierarchical models with implicit $q(\psi \mid x)$ and suggested a multisample ELBO surrogate to be used in optimization. \citet{molchanov2018doubly} have shown that the proposed surrogate is actually a lower bound on ELBO, and effectively gives a novel lower bound on the entropy of the marginal $q(z \mid x)$ and extended the SIVI to the case of semi-implicit priors, proposing Doubly Semi-Implicit Variational Inference (DSIVI). They have also shown both theoretically and empirically that performing inference in the "marginal" space of $z$ as opposed to the joint space $z, \psi$ is beneficial in cases when the decoder $p(x \mid z)$ does not use $\psi$ directly.

The core contribution of the paper is a novel upper bound on marginal log-likelihood. Previously, \citet{dieng2017variational,kuleshov2017neural} suggested using $\chi^2$-divergence to give a variational upper bound to the marginal log-likelihood. However, their bound was not an expectation of a random variable, but instead a logarithm of the expectation, preventing unbiased stochastic optimization. \citet{jebara2001reversing} reverse Jensen's inequality to give a variational upper bound in case of mixtures of exponential family distributions by extensive use of the problem's structure. Related to our core idea of joint sampling $z$ and $\psi_0$ in \eqref{eq:kl-upper-bound} is an observation of \citet{Grosse2015SandwichingTM} that Annealed Importance Sampling (AIS, \citet{neal2001annealed}) ran backward from the auxiliary variable sample $\psi_0$ gives an unbiased estimate of $1/q(z \mid x)$, and thus can also be used to upper bound the marginal log-density. However, AIS-based estimation is too computationally expensive to be used during training.

\section{Experiments}\label{sec:exps}
We only focus on cases with explicit prior $p(z)$ to simplify comparison to prior work. Hierarchical priors correspond to nested variational inference, to which most of the variational inference results readily apply \citep{atanov2018the}.

\begin{figure}
  \centering
  \begin{subfigure}[b]{0.49 \textwidth}
    \includegraphics[width=\textwidth]{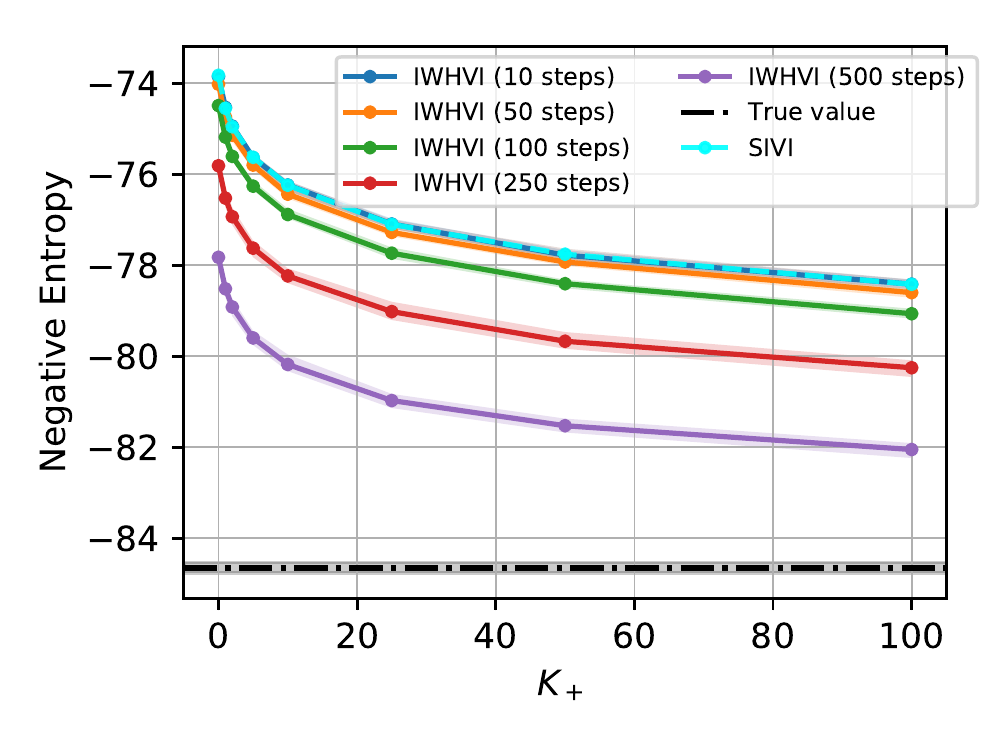}
    \caption{Negative entropy bounds for 50-dimensional Laplace distribution. Shaded area denotes 90\% confidence interval computed over 50 independent runs for each $K$.}
    \label{fig:toy-laplace}
  \end{subfigure}
  \hfill
  \begin{subfigure}[b]{0.49 \textwidth}
    \includegraphics[width=\textwidth]{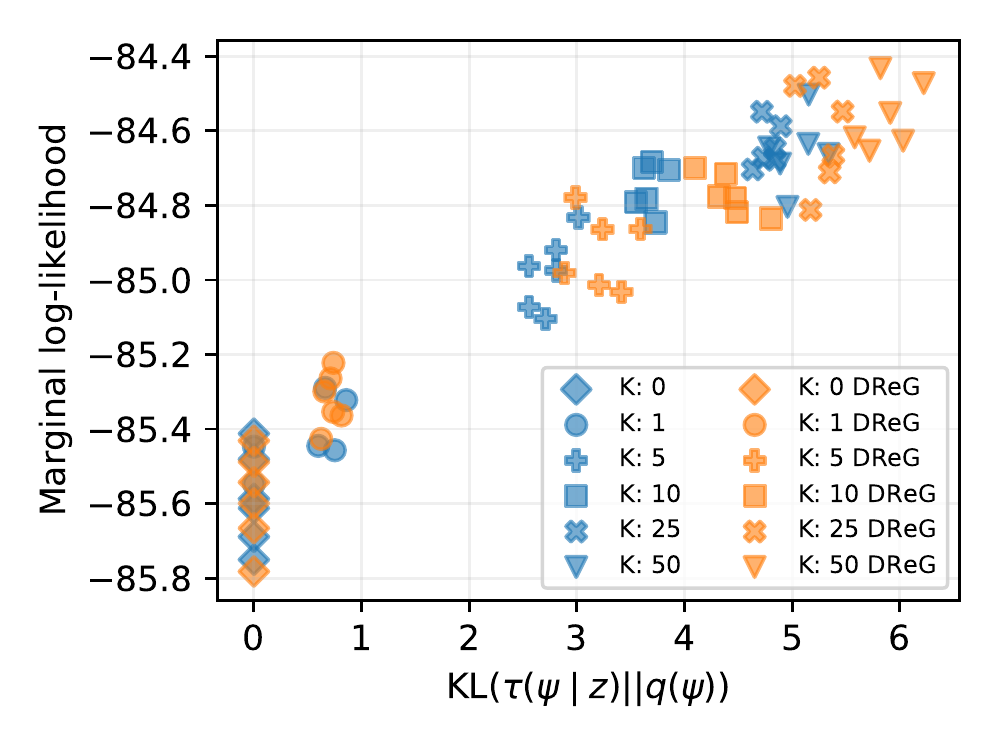}
    \caption{Final marginal log-likelihood $\log p(x)$ estimates and expected $D_{KL}(\tau(\psi \mid z) \mid\mid q(\psi))$ for IWHVI-based VAEs trained with different $K$. Each model was trained and plotted 6 times.}
    \label{fig:k_study}
  \end{subfigure}
\end{figure}

\subsection{Toy Experiment} \label{sec:toy-exp}
As a toy experiment we consider a 50-dimensional factorized standard Laplace distribution $q(z)$ as a hierarchical scale-mixture model:
\begin{align*}
q(z)
= \prod_{d=1}^{50} \text{Laplace}(z_d \mid 0, 1)
= \int \prod_{d=1}^{50} \mathcal{N}(z_d \mid 0, \psi_d) \text{Exp}(\psi_d \mid \tfrac{1}{2}) d\psi_{1:50}
\end{align*}
We do not make use of factorized joint distribution $q(z, \psi)$ to explore bound's behavior in high dimensions. We use the proposed bound from \Cref{thm:upperboundthm} and compare it to SIVI \citep{yin2018semi} on the task of upper-bounding the negative differential entropy $\E_{q(z)} \log q(z)$. For IWHVI we take $\tau(\psi \mid z)$ to be a Gamma distribution whose concentration and rate are generated by a neural network with three 500-dimensional hidden layers from $z$. We use the freedom to design architecture and initialize the network at prior. Namely, we also add a sigmoid "gate" output with large initial negative bias and use the gate to combine prior concentration and rate with those generated by the network. This way we are guaranteed to perform no worse than SIVI even at a randomly initialized $\tau$. \Cref{fig:toy-laplace} shows the value of the bound for a different number of optimization steps over $\tau$ parameters, minimizing the bound. The whole process (including random initialization of neural networks) was repeated 50 times to compute empirical 90\% confidence intervals. As results clearly indicate, the proposed bound can be made much tighter, more than halving the gap to the true negative entropy.

\subsection{Variational Autoencoder} \label{sec:vae-exp}
\begin{figure}
  \begin{minipage}[c]{0.55 \textwidth}
  \vspace{-2em}
\begin{tabularx}{\linewidth}{lcc}
\hline
\textbf{Method}   & \textbf{MNIST}                             & \textbf{OMNIGLOT}   \\ \hline
\multicolumn{3}{c}{\small From \citep{mescheder2017adversarial}}       \\
    AVB + AC          & \multicolumn{1}{c}{$-83.7 \pm 0.3$} &  --- \\
    \cline{1-3}
    \textbf{IWHVI}    & \multicolumn{1}{c|}{$-83.9 \pm 0.1$} & $-104.8 \pm 0.1$  \\
    SIVI              & \multicolumn{1}{c|}{$-84.4 \pm 0.1$} & $-105.7 \pm 0.1$  \\
    HVM               & \multicolumn{1}{c|}{$-84.9 \pm 0.1$} & $-105.8 \pm 0.1$     \\
    VAE + RealNVP     & \multicolumn{1}{c|}{$-84.8 \pm 0.1$} & $-106.0 \pm 0.1$  \\
    VAE + IAF         & \multicolumn{1}{c|}{$-84.9 \pm 0.1$} & $-107.0 \pm 0.1$ \\
    VAE               & \multicolumn{1}{c|}{$-85.0 \pm 0.1$} & $-106.6 \pm 0.1$ \\
\end{tabularx}
  \end{minipage}
  \hfill
\begin{minipage}[c]{0.45 \textwidth}
    \includegraphics[width=\linewidth]{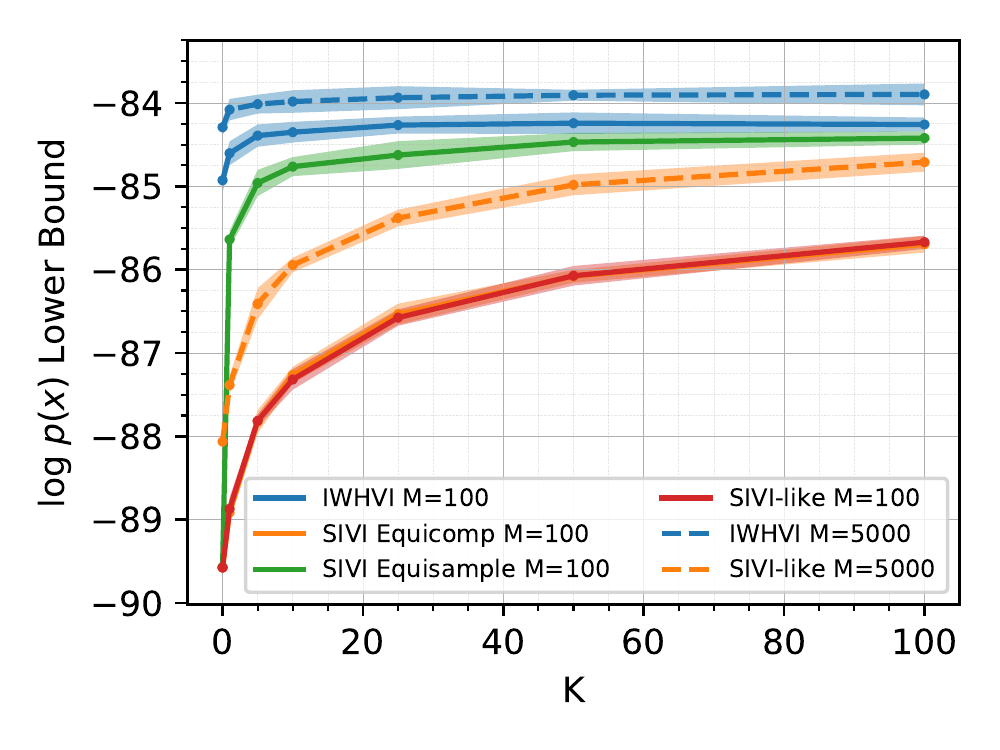}
  \end{minipage}
  \vspace{-1em}
   \caption{\textbf{Left}: Test log-likelihood on dynamically binarized MNIST and OMNIGLOT. \textbf{Right}: Comparison of multisample DIWHVI and SIVI-IW on a trained MNIST VAE from \cref{sec:vae-exp} for $M = 100$ and $5000$. Shaded area denotes $\pm 2$ std. interval, computed over 10 independent runs for each value of $K$.}
   \label{tbl:iwhvae}
   \label{fig:vae-eval-gap}
\end{figure}
We further test our method on the task of generative modeling, applying it to VAE \citep{kingma2013auto}, which is a standard benchmark for inference methods. Ideally, better inference should allow one to learn more expressive generative models. We report results on two datasets: MNIST \citep{lecun1998gradient} and OMNIGLOT \citep{lake2015human}. For MNIST we follow the setup by \citet{mescheder2017adversarial}, and for OMNIGLOT we follow the standard setup \citep{burda2015importance}. For experiment details see \cref{sec:vae-exp-details}.

During training we used the proposed bound \cref{eq:bound} with analytically tractable prior $p(z) = \mathcal{N}(z \mid 0, 1)$ with increasing number $K$: we used $K = 0$ for the first 250 epochs, $K=5$ for the next 250 epochs, and $K = 25$ for the next 500 epochs, and $K=50$ from then on. We used $M = 1$ during training.

To estimate the marginal log-likelihood for hierarchical models (IWHVI, SIVI, HVM) we use the DIWHVI lower bound \eqref{eq:iw-bound} for $M=5000$, $K=100$ (for justification of DIWHVI as an evaluation metric see \cref{sec:iwhvi-eval-exp}). Results are shown in \cref{tbl:iwhvae}. To evaluate the SIVI using the DIWHVI bound we fit $\tau$ to a trained model by making 7000 epochs on the trainset with $K=50$, keeping parameters of $q_\phi(z, \psi \mid x)$ and $p_\theta(x, z)$ fixed. We observed improved performance compared to special cases of HVM and SIVI, and the method showed comparable results to the prior works.

For HVM on MNIST we observed its $\tau(\psi \mid z)$ essentially collapsed to $q(\psi)$, having expected KL divergence between the two extremely close to zero. This indicates the "posterior collapse" \citep{pmlr-v80-kim18e,chen2016variational} problem where the inference network $q(z \mid \psi)$ chose to ignore the extra input $\psi$ and effectively degenerated to a vanilla VAE. At the same time IWHVI does not suffer from this problem due to non-zero $K$, achieving average $D_{KL}({\tau(\psi \mid z, x)} \mid\mid {q(\psi)})$ of approximately 6.2 nats, see \cref{sec:iwhvi-eval-exp}. For OMNIGLOT HVM did learn useful $\tau$, and achieved average $D_{KL}({\tau(\psi \mid z, x)} \mid\mid {q(\psi)}) \approx 1.98$ nats, however IWHVI did much better and achieved $\approx 9.97$ nats.

To investigate $K$'s influence on the training process, we trained VAEs on MNIST for 3000 epochs for different values of $K$ and evaluated DIWHVI bound for $M = 1000, K = 100$ and plotted results in \cref{fig:k_study}. One can see that higher values of $K$ lead to better final models in terms of marginal log-likelihood, as well as more informative auxiliary inference networks $\tau(\psi \mid z, x)$ compared to the prior $q(\psi)$. Using IWHVI-DReG gradient estimator (see \cref{sec:snr-details}) increased the KL divergence, but resulted in only a modest increase in the marginal log-likelihood.

\subsection{DIWHVI as Evaluation Metric} \label{sec:iwhvi-eval-exp}
One of the established approaches to evaluate the intractable marginal log-likelihood in Latent Variable Models is to compute the multisample IWAE-bound with large $M$ since it is shown to converge to the marginal log-likelihood as $M$ goes to infinity. Since both IWHVI and SIVI allow tightening the bound by taking more samples $z_m$, we compare methods along this direction.

Both DIWHVI and SIVI (being a special case of the former) can be shown to converge to marginal log-likelihood as both $M$ and $K$ go to infinity, however, rates might differ. We empirically compare the two by evaluating an MNIST-trained IWHVAE model from \cref{sec:vae-exp} for several different values $K$ and $M$. We use the proposed DIWHVI bound \eqref{eq:iw-bound}, and compare it with several SIVI modifications. We call \emph{SIVI-like} the \eqref{eq:iw-bound} with $\tau(\psi \mid z) = q(\psi)$, but without $\psi$ reuse, thus using $M K$ independent samples. \emph{SIVI Equicomp} stands for sample reusing bound \eqref{eq:sivi-reused}, which uses only $M + K$ samples, and uses same $\psi_{1:K}$ for every $z_m$. \emph{SIVI Equisample} is a fair comparison in terms of the number of samples: we take $M (K + 1)$ samples of $\psi$, and reuse $M K$ of them for every $z_m$. This way we use the same number of samples $\psi$ as DIWHVI does, but perform $O(M^2 K)$ log-density evaluations to estimate $\log q(z \mid x)$, which is why we only examine the $M = 100$ case.

Results shown in \cref{fig:vae-eval-gap} indicate superior performance of the DIWHVI bound. Surprisingly SIVI-like and SIVI Equicomp estimates nearly coincide, with no significant difference in variance; thus we conclude sample reuse does not hurt SIVI. Still, there is a considerable gap to the IWHVI bound, which uses similar to SIVI-like amount of computing and samples. In a more fair comparison to the Equisample SIVI bound, the gap is significantly reduced, yet IWHVI is still a superior bound, especially in terms of computational efficiency, as there are no $O(M^2 K)$ operations.

Comparing IWHVI and SIVI-like for $M=5000$ we see that the former converges after a few dozen samples, while SIVI is rapidly improving, yet lagging almost 1 nat behind for 100 samples, and even 0.5 nats behind the HVM bound (IWHVI for $K=0$). One explanation for the observed behaviour is large $\mathbb{E}_{q(z \mid x)} D_{KL}(q(\psi \mid x) \mid\mid q(\psi \mid x, z))$, which was estimated \footnote{Difference between $K$-sample IWAE and ELBO gives a lower bound on $D_{KL}(\tau(\psi \mid z) \mid\mid q(\psi \mid z))$, we used $K = 5000$.} (on a test set) to be at least $46.85$ nats, causing many samples from $q(\psi)$ to generate poor likelihood $q(z \mid \psi)$ for a given $z_m$ due to large difference with the true inverse model $q(\psi \mid x, z)$. This is consistent with motivation layed out in \cref{sec:sivi-insights}: a better approximate inverse model $\tau$ leads to more efficient sample usage. At the same time $\mathbb{E}_{q(z \mid x)} D_{KL}(\tau(\psi \mid x, z) \mid\mid q(\psi \mid x, z))$ was estimated to be approximately $3.24$ and $\mathbb{E}_{q(z \mid x)} D_{KL}(\tau(\psi \mid x, z) \mid\mid q(\psi \mid x)) \approx 6.25$, proving that one can indeed do much better by learning $\tau(\psi \mid x, z)$ instead of using the prior $q(\psi \mid x)$.

\section{Conclusion}
We presented a variational upper bound on the marginal log density, which allowed us to formulate sandwich bounds for $D_{KL}(q(z \mid x) \mid\mid p(z))$ for the case of hierarchical model $q(z \mid x)$ in addition to prior works that only provided multisample variational upper bounds for the case of $p(z$ being a hierarchical model. We applied it to lower bound the intractable ELBO with a tractable one for the case of latent variable model approximate posterior $q_\phi(z \mid x)$. We experimentally validated the bound and showed it alleviates regularizational effect to a further extent than prior works do, allowing for more expressive approximate posteriors, which does translate into a better inference. We then combined our bound with multisample IWAE bound, which led to a tighter lower bound of the marginal log-likelihood. We therefore believe the proposed variational inference method will be useful for many variational models.

\section*{Acknowledgements}
Authors would like to thank Aibek Alanov, Dmitry Molchanov and Oleg Ivanov for valuable discussions and feedback.

\bibliography{iwhvi}

\newpage
\appendix

% \icmltitle{Supplementary Materials for \\ Importance Weighted Hierarchical Variational Inference}

\section{Proofs}

\begin{theorem}[Marginal log-density upper bound] \label{thm:upperboundthm}
For any $q(z, \psi)$, $K \in \mathbb{N}_0$ and $\tau(\psi \mid z)$ such that for any $\psi$ s.t. $\tau(\psi \mid z) = 0$ we have $q(\psi, z) = 0$, consider
$$
\mathcal{U}_{K} = \E_{q(\psi_0 \mid z)} \E_{\tau(\psi_{1:K} \mid z)} \log \left( \frac{1}{K+1} \sum_{k=0}^{K} \frac{q(z, \psi_k)}{\tau(\psi_k \mid z)} \right)
$$
where we write $\tau(\psi_{1:K} \mid z) = \prod_{k=1}^{K} \tau(\psi_k \mid z)$ for brevity. Then the following holds:
\begin{enumerate}
    \item $\mathcal{U}_{K} \ge \log q(z)$
    \item $\mathcal{U}_{K} \ge \mathcal{U}_{K + 1}$
    \item $\lim\limits_{K \to \infty} \mathcal{U}_K = \log q(z)$.
\end{enumerate}
\begin{proof}

\begin{enumerate}
    \item Consider a gap between the proposed bound at the marginal log density:

\begin{align*}
\mbox{Gap}
&= 
\E_{q(\psi_0 \mid z)} \E_{\tau(\psi_{1:K} \mid z)} \log \left( \frac{1}{K+1} \sum_{k=0}^{K} \frac{q(z, \psi_k)}{\tau(\psi_k \mid z)} \right) - \log q(z) \\
&= 
\E_{q(\psi_0 \mid z)} \E_{\tau(\psi_{1:K} \mid z)} \log \left( \frac{1}{K+1} \sum_{k=0}^{K} \frac{q(\psi_k \mid z)}{\tau(\psi_k \mid z)} \right) \\
& = 
\E_{q(\psi_0 \mid z)} \E_{\tau(\psi_{1:K} \mid z)} \log \left( \frac{q(\psi_0 \mid z) \tau(\psi_{1:K} \mid z)}{\omega_{q,\tau}(\psi_{0:K} \mid z)} \right) \\
& = 
D_{KL}(q(\psi_0 \mid z) \tau(\psi_{1:K} \mid z) \mid\mid \omega_{q,\tau}(\psi_{0:K} \mid z))
\ge 0
\end{align*}

Where the last line holds due to $\omega_q$ being a normalized density function (see \Cref{lem:omega_dist}):
$$
\omega_{q,\tau}(\psi_{0:K} \mid z)
=
\frac{q(\psi_0 \mid z) \tau(\psi_{1:K} \mid z)}{\frac{1}{K+1} \sum_{k=0}^{K} \frac{q(\psi_k \mid z)}{\tau(\psi_k \mid z)}}
$$

\item Now we will prove the second claim.

\begin{align*}
\mathcal{U}_{K} - \mathcal{U}_{K + 1}
&= \E_{q(\psi_0 \mid z)} \E_{\tau(\psi_{1:K+1} \mid z)} \log \frac{\frac{1}{K+1} \sum_{k=0}^{K} \frac{q(z, \psi_k)}{\tau(\psi_k \mid z)}}{\frac{1}{K+2} \sum_{k=0}^{K+1} \frac{q(z, \psi_k)}{\tau(\psi_k \mid z)}} \\
&= \E_{q(\psi_0 \mid z)} \E_{\tau(\psi_{1:K+1} \mid z)} \log \frac{q(\psi_0 \mid z) \tau(\psi_{1:K+1} \mid z)}{\nu_{q,\tau}(\psi_{0:K+1} \mid z)} \\
&= D_{KL} \left( q(\psi_0 \mid z) \tau(\psi_{1:K+1} \mid z) \mid\mid \nu_{q,\tau}(\psi_{0:K+1} \mid z) \right) \ge 0
\end{align*}

Where we used the fact that $\nu_{q,\tau}(\psi_{0:K+1} \mid z)$ is normalized density due to  \Cref{lem:nu_dist}
$$
\nu_{q,\tau}(\psi_{0:K+1} \mid z) = \omega_{q,\tau}(\psi_{0:K} \mid z) \tau(\psi_{K+1} \mid z) \frac{1}{K+2} \sum_{k=0}^{K+1} \frac{q(\psi_k \mid z)}{\tau(\psi_k \mid z)}
$$

\item For the last claim we follow \citep{burda2015importance}. Consider
$$
M_{K}
= \frac{1}{K+1} \sum_{k=0}^{K} \frac{q(z, \psi_k)}{\tau(\psi_k \mid z)}
= \overbrace{\frac{1}{K+1} \frac{q(z, \psi_0)}{\tau(\psi_0 \mid z)}}^{A_{K}} + \overbrace{\frac{K}{K+1}}^{B_{K}} \overbrace{\frac{1}{K} \sum_{k=1}^{K} \frac{q(z, \psi_k)}{\tau(\psi_k \mid z)}}^{X_{K}}
$$
Due to Law of Large Numbers we have
$$
A_{K} \xrightarrow[K\to\infty]{a.s.} 0,
\quad\quad
X_{K} \xrightarrow[K\to\infty]{a.s.} \E_{\tau(\psi \mid z)} \frac{q(z, \psi)}{\tau(\psi \mid z)} = q(z),
\quad\quad
B_{K} \xrightarrow[K\to\infty]{a.s.} 1
$$

Thus
$$
M_{K} \xrightarrow[K\to\infty]{a.s.} q(z),
\quad\quad
\mathcal{U}_{K} = \E_{\tau(\psi_{0:K} \mid z)} \log M_{K} \xrightarrow[K \to \infty]{} \log q(z)
$$
\end{enumerate}

\end{proof}
\end{theorem}

\begin{lemma}[$\omega_{q,\tau}$ distribution, following \citet{domke2018importance}]\label{lem:omega_dist}
Given $z$, consider a following generative process:
\begin{itemize}
    \item Sample $K+1$ i.i.d. samples from $\hat\psi_k \sim \tau(\psi \mid z)$
    \item For each sample compute its weight $w_k = \frac{q(\hat\psi_k, z)}{\tau(\hat\psi_k \mid z)}$
    \item Sample $h \sim \text{Cat}\left(\frac{w_0}{\sum_{k=0}^{K} w_k}, \dots, \frac{w_{K}}{\sum_{k=0}^{K} w_k}\right)$
    \item Put $h$-th sample first, and then the rest: $\psi_0 = \hat\psi_h$, $\psi_{1:K} = \hat\psi_{\backslash h}$
\end{itemize}
Then the marginal density of $\psi_{0:K}$
$$
\omega_{q,\tau}(\psi_{0:K} \mid z)
=
\frac{q(\psi_0 \mid z) \tau(\psi_{1:K} \mid z)}{\frac{1}{K+1} \sum_{k=0}^{K} \frac{q(\psi_k \mid z)}{\tau(\psi_k \mid z)}}
$$
\begin{proof}
The joint density for the generative process described above is
$$
\omega_{q,\tau}(\hat\psi_{0:K}, h, \psi_{0:K} \mid z)
=
\tau(\hat\psi_{0:K} \mid z)
\tfrac{w_h}{\sum_{k=0}^{K} w_k}
\delta(\psi_0 - \hat\psi_h) \delta(\psi_{1:K} - \hat\psi_{\backslash h})
$$

One can see that this is indeed a normalized density
\begin{align*}
\int \sum_{h=0}^{K}
&\left(
\int
\omega_\tau(\hat\psi_{0:K}, h, \psi_{0:K} \mid z)
d\psi_{0:K}
\right)
d\hat\psi_{0:K}
=
\int \sum_{h=0}^{K}
\tau(\hat\psi_{0:K} \mid z)
\tfrac{w_h}{\sum_{k=0}^{K} w_k}
d\hat\psi_{0:K} \\
& =
\int
\tau(\hat\psi_{0:K} \mid z)
\sum_{h=0}^{K} \tfrac{w_h}{\sum_{k=0}^{K} w_k}
d\hat\psi_{0:K}
 =
\int
\tau(\hat\psi_{0:K} \mid z)
d\hat\psi_{0:K}
= 1
\end{align*}

The marginal density $\omega_{q,\tau}(\psi_{0:K} \mid z)$ then is

\begin{align*}
\omega_{q,\tau}(\psi_{0:K} \mid z)
& =
\int
\sum_{h=0}^{K}
\tau(\hat\psi_{0:K} \mid z)
\tfrac{w_h}{\sum_{k=0}^{K} w_k}
\delta(\psi_0 - \hat\psi_h) \delta(\psi_{1:K} - \hat\psi_{\backslash h})
d \hat\psi_{0:K} \\
& =
(K + 1) \int
\tau(\hat\psi_{0:K} \mid z)
\tfrac{w_0}{\sum_{k=0}^{K} w_k}
\delta(\psi_0 - \hat\psi_0) \delta(\psi_{1:K} - \hat\psi_{1:K})
d \hat\psi_{0:K} \\
& =
\int
\tau(\hat\psi_{1:K} \mid z)
\tfrac{q(z, \hat\psi_0)}{\tfrac{1}{K + 1} \sum_{k=0}^{K} w_k}
\delta(\psi_0 - \hat\psi_0) \delta(\psi_{1:K} - \hat\psi_{1:K})
d \hat\psi_{0:K} \\
& =
\tau(\psi_{1:K} \mid z)
\tfrac{q(z, \psi_0)}{\tfrac{1}{K + 1} \sum_{k=0}^{K} \tfrac{q(\psi_k, z)}{\tau(\psi_k \mid z)}}
 =
\tfrac{q(\psi_0 \mid z) \tau(\psi_{1:K} \mid z)}{\tfrac{1}{K + 1} \sum_{k=0}^{K} \tfrac{q(\psi_k \mid z)}{\tau(\psi_k \mid z)}}
\end{align*}

Where on the second line we used the fact that integrand is symmetric under the choice of $h$.

\end{proof}
\end{lemma}

\begin{lemma}\label{lem:nu_dist}
Let
$$
\nu_{q,\tau}(\psi_{0:K+1} \mid z) = \omega_{q,\tau}(\psi_{0:K} \mid z) \tau(\psi_{K+1} \mid z) \frac{1}{K+2} \sum_{k=0}^{K+1} \frac{q(\psi_k \mid z)}{\tau(\psi_k \mid z)}
$$

Then $\nu_{q,\tau}(\psi_{0:K+1} \mid z)$ is a normalized density.
\begin{proof}
$\nu_{q,\tau}(\psi_{0:K+1} \mid z)$ is non-negative due to all the terms being non-negative. Now we'll show it integrates to 1 (colors denote corresponding terms):

\begin{align*}
\int &\omega_{q,\tau}(\psi_{0:K} \mid z) {\color{blue}\tau(\psi_{K+1} \mid z)} {\color{red} \frac{1}{K+2} \sum_{k=0}^{K+1} \frac{q(\psi_k \mid z)}{\tau(\psi_k \mid z)}} d\psi_{0:K+1} \\
& = {\color{red} \frac{1}{K+2}} {\color{violet} \int \omega_{q, \tau}(\psi_{0:K} \mid z)} \left[{ \color{red} \sum_{k=0}^K \frac{q(\psi_k \mid z)}{\tau(\psi_k \mid z)} + {\color{blue} \int \tau(\psi_{K+1} \mid z)} \frac{q(\psi_{K+1} \mid z)}{\tau(\psi_{K+1} \mid z)} } { \color{blue} d\psi_{K+1} } \right] d\psi_{0:K}  \\
& = \frac{1}{K+2} \left[\int {\color{violet} \frac{q(\psi_0 \mid z) \tau(\psi_{1:K} \mid z)}{\frac{1}{\color{orange} K+1} \sum_{k=0}^{K} \frac{q(\psi_k \mid z)}{\tau(\psi_k \mid z)}}} \sum_{k=0}^K \frac{q(\psi_k \mid z)}{\tau(\psi_k \mid z)} d\psi_{0:K} + 1\right] = \frac{{\color{orange} K+1} + 1}{K+2} = 1 \\
\end{align*}
\end{proof}
\end{lemma}

\begin{theorem}[DIWHVI Evidence Lower Bound] \label{th:iwae-bound}
\begin{align}
\log p(x)
\ge
\E
\log 
\left[\frac{1}{M} \sum_{m=1}^M \frac{p(x \mid z_m) \frac{1}{K} \sum_{k=1}^{K} \frac{p(z_m, \zeta_{m,k})}{\rho(\zeta_{m,k} \mid z_m)}}{\frac{1}{K + 1} \sum_{k=0}^{K} \frac{q(z_m, \psi_{m,k} \mid x)}{\tau(\psi_{m,k} \mid z_m, x)}}
\right]
\end{align}

Where the expectation is taken over the following generative process:
\begin{enumerate}
    \item Sample $\psi_{m,0} \sim q(\psi \mid x)$ for $1 \le m \le M$
    \item Sample $z_{m} \sim q(z \mid x_n, \psi_{m,0})$ for $1 \le m \le M$
    \item Sample $\psi_{m,k} \sim \tau(\psi \mid z_{m}, x)$ for $1 \le m \le M$ and $1 \le k \le K$
    \item Sample $\zeta_{m,k} \sim \rho(\zeta \mid z_{m})$ for $1 \le m \le M$ and $1 \le k \le K$
\end{enumerate}

\begin{proof}
Consider a random variable
$$
X_M = \frac{1}{M} \sum_{m=1}^M \frac{p(x \mid z_m) \frac{1}{K} \sum_{k=1}^{K} \frac{p(z_m, \zeta_{m,k})}{\rho(\zeta_{m,k} \mid z_m)}}{\frac{1}{K + 1} \sum_{k=0}^{K} \frac{q(z_m, \psi_{m,k} \mid x)}{\tau(\psi_{m,k} \mid z_m, x)}}
$$

\begin{fleqn}%[0.5em]
We'll show it's an unbiased estimate of $p(x)$ (colors denote corresponding terms) and then just invoke Jensen's inequality:
\begin{align*}
\E X_M
& = \int
\Biggl[
 \left( \prod_{m=1}^M q(\psi_{m,0} \mid x) q(z_m \mid \psi_{m,0}, x)
    \tau(\psi_{m,1:K} \mid z_m, x) {\color{red} \rho(\zeta_{m,1:K} \mid z_m)} \right) \\
& \quad\quad\quad \frac{1}{M} \sum_{m=1}^M \frac{p(x \mid z_m) {\color{red} \frac{1}{K} \sum_{k=1}^{K} \frac{p(z_m, \zeta_{m,k})}{\rho(\zeta_{m,k} \mid z_m)}}}{\frac{1}{K + 1} \sum_{k=0}^{K} \frac{q(z_m, \psi_{m,k} \mid x)}{\tau(\psi_{m,k} \mid z_m, x)}} \Biggr] d\psi_{1:M,0:K} d\zeta_{1:M,1:K} dz_{1:M} \\
\end{align*}
First, we move in the integral w.r.t. $\zeta$ into the numerator:
\begin{align*}
\phantom{\E X_M} = \int \Biggl(
\frac{1}{M} \sum_{m=1}^M & \frac{p(x \mid z_m) {\color{red} \E\limits_{\rho(\zeta_{m,1:K} \mid z_m)} \frac{1}{K} \sum\limits_{k=1}^{K} \frac{p(z_m, \zeta_{m,k})}{\rho(\zeta_{m,k} \mid z_m)}} }{\frac{1}{K + 1} \sum\limits_{k=0}^{K} \frac{q(z_m, \psi_{m,k} \mid x)}{\tau(\psi_{m,k} \mid z_m, x)}} 
\\
&\prod_{m=1}^M q(\psi_{m,0} \mid x) q(z_m \mid \psi_{m,0}, x)
    \tau(\psi_{m,1:K} \mid z_m, x)
d\psi \Biggr) dz \\
\end{align*}
Next, we leverage $z_m$'s independence:
\begin{align*}
\phantom{\E X_M}
& = \int
\Biggl[
 \left( {\color{blue} \prod_{m=1}^M } q(\psi_{m,0} \mid x) q(z_m \mid \psi_{m,0}, x)
    \tau(\psi_{m,1:K} \mid z_m, x) \right)
\\ & \quad\quad\quad
 {\color{blue} \frac{1}{M} \sum_{m=1}^M} \frac{p(x \mid z_m) {\color{red} p(z_m)}}{\frac{1}{K + 1} \sum_{k=0}^{K} \frac{q(z_m, \psi_{m,k} \mid x)}{\tau(\psi_{m,k} \mid z_m, x)}} \Biggr] d\psi_{1:M,0:K} dz_{1:M}\\
& = 
{\color{blue} \frac{1}{M} \sum_{m=1}^M} 
\int
p(x, z_m) 
{\color{violet} \frac{q(z_m, \psi_{m,0} \mid x) \tau(\psi_{m,1:K} \mid z_m)}{\frac{1}{K + 1} \sum_{k=0}^{K} \frac{q(z_m, \psi_{m,k} \mid x)}{\tau(\psi_{m,k} \mid z_m, x)}}} d\psi_{m,0:K} dz_{m}\\
& = 
\frac{1}{M} \sum_{m=1}^M 
\int
p(x, z_m) 
{\color{violet} \omega_{q,\tau}(\psi_{m,0:K} \mid z_m, x)} d\psi_{m,0:K} dz_{m} \\
&= 
\frac{1}{M} \sum_{m=1}^M 
\int p(x, z_m) dz_{m}  = p(x)
\end{align*}

\end{fleqn}

Where $\omega_{q,\tau}(\psi_{m,0:K} \mid z_m, x)$ is a density from the \Cref{lem:omega_dist}. Now the rest follows from the Jensen's inequality due to logarithm's concavity:

$$
\log p(x) = \log \E X_M \ge \E \log X_M
$$
\end{proof}
\end{theorem}

\begin{corollary}
All statements of Theorem 1 of \citep{burda2015importance} apply to this bounds as well.
\end{corollary}

%\section{Putting Prior Work in Perspective}
%In this section we systematize how previous methods look through the lenses of the proposed bound.
%
%\begin{figure}[H]
%\centering
%\begin{tabular}{p{5cm}|c|c|c|c}
%\multicolumn{1}{c}{\textbf{Method}} & \multicolumn{1}{c}{\textbf{$p(z)$ estimate}} & \multicolumn{1}{c}{\textbf{$\zeta_k$ distribution}} & \multicolumn{1}{c}{\textbf{$q(z)$ estimate}} & \multicolumn{1}{c}{\textbf{$\psi_k$ distribution}} \\
%\hline
%\textbf{DSIVI} \citep{molchanov2018doubly} & $ \frac{1}{K} \sum\limits_{k=1}^K p(z \mid \zeta_k)$ & $p(\zeta)$ & $\frac{1}{K+1} \sum\limits_{k=0}^K q(z \mid \psi_k, x)$ & $q(\psi \mid x)$ \\
%\textbf{SIVI} \citep{yin2018semi} & explicit $p(z)$ & --- & $\frac{1}{K+1} \sum\limits_{k=0}^K q(z \mid \psi_k, x)$ & $q(\psi \mid x)$ \\
%\textbf{HVM} \citep{ranganath2016hierarchical}, \textbf{Auxiliary Variables} \citep{agakov2004auxiliary} & explicit $p(z)$ & --- & $\frac{q(z, \psi \mid x)}{\tau(\psi \mid x, z)}$ & --- \\
%\textbf{Joint Bound} & explicit $p(z)$ & --- & $\frac{q(z, \psi \mid x)}{p(\psi \mid z)}$ & --- \\
%\end{tabular}
%\end{figure}

\section{Special Cases} \label{sec:special-cases}
Many previously known methods can be seen as special cases of the proposed bound. In particular:
\begin{itemize}
    \item For an arbitrary $K,L$, $\tau(\psi \mid z, x) = q(\psi \mid x)$ (the hyperprior on $\psi$ under $q$) and $\rho(\zeta \mid z) = p(\zeta)$ we recover the DSIVI bound \citep{molchanov2018doubly}
    $$ \log p(x) \ge \E_{q(z, \psi_0 \mid x)} \E_{q(\psi_{1:K} \mid x)} \E_{p(\zeta_{1:K})} \log \frac{p(x \mid z) \tfrac{1}{L} \sum_{l=1}^{L} p(z \mid \zeta_l)}{\tfrac{1}{K + 1} \sum_{k=0}^{K} q(z \mid \psi_k, x)} $$
    \item For an arbitrary $K$, $\tau(\psi \mid z, x) = q(\psi \mid x)$ and an explicit prior $p(z)$ (equivalently, $\rho(\zeta \mid z) = p(\zeta \mid z)$) we recover the SIVI bound \citep{yin2018semi} $$ \log p(x) \ge \E_{q(z, \psi_0 \mid x)} \E_{q(\psi_{1:K} \mid x)} \log \frac{p(x, z)}{\tfrac{1}{K + 1} \sum_{k=0}^K q(z | \psi_k, x)} $$
    \item For $K = 0$, arbitrary $\tau(\psi \mid z, x)$ and explicit prior $p(z)$ (equivalently, $\rho(\zeta \mid z) = p(\zeta \mid z)$) we recover the HVM bound \citep{ranganath2016hierarchical}, also known as auxiliary variables bound \citep{agakov2004auxiliary,salimans2015markov,pmlr-v48-maaloe16} $$ \log p(x) \ge \E_{q(z, \psi_0 \mid x)} \log \frac{p(x, z)}{\tfrac{q(z, \psi_0 \mid x)}{\tau(\psi_0 \mid z, x)}} $$
    \item For $K = 0$, $\tau(\psi \mid z, x) = p(\psi \mid z, x)$ and structurally similar to $q(z \mid x)$ prior $p(z) = \int p(z, \psi) d\psi$ we recover the joint bound \citep{NIPS2017_6921} $$ \log p(x) \ge \E_{q(z, \psi_0 \mid x)} \log \frac{p(x \mid z) p( z, \psi_0)}{q(z, \psi_0 \mid x)} $$
    \item For an arbitrary $K$, factorized inference and prior models $q(z, \psi \mid x) = q(z \mid x) q(\psi \mid x)$, $p(z, \zeta) = p(z) p(\zeta)$, optimal $\tau(\psi \mid z, x) = q(\psi \mid x)$ and $\rho(\zeta \mid z) = p(\zeta)$ we recover the standard ELBO $$ \log p(x) \ge \E_{q(z \mid x)} \log \frac{p(x, z)}{q(z \mid x)} $$ So even if there's no hierarchical structure, the bound still works.
\end{itemize}

\section{Lower Bound on KL divergence between hierarchical models} \label{sec:kl-lower-bound}
Similarly to \eqref{eq:kl-upper-bound} we can give a lower bound on KL divergence:
\begin{align} 
D_{KL}(q(z \mid x) \mid\mid \; p(z))
&\ge
\E_{q(z \mid x)}
\E_{\tau(\psi_{1:K} \mid x, z)}
\E_{p(\zeta_0 \mid z)} 
\E_{\rho(\zeta_{1:L} \mid z)} 
\Biggl[
\log
\frac{
    \tfrac{1}{K} \sum_{k=1}^{K} \tfrac{q(z, \psi_k \mid x)}{\tau(\psi_k \mid x, z)}
}{
    \tfrac{1}{L + 1} \sum_{l=0}^{L} \frac{p(z, \zeta_l)}{\rho(\zeta_l \mid z)}
}
\Biggr]
\label{eq:kl-lower-bound}
\end{align}
Unfortunately, this lower bound requires sampling from the true inverse $p(\zeta|z)$ and does not allow the same trick as \eqref{eq:kl-upper-bound}, leaving us with expensive posterior sampling techniques. However, unlike the previously suggested Fenchel conjugate based lower bound \citep{NIPS2016_6066,molchanov2018doubly}, the bound \cref{eq:kl-lower-bound} not only uses samples from $z$, but also makes use of its underlying density and can be made increasingly tighter by increasing $K$.

Analyzing the gaps between the marginal log-density $\log q_\phi(z \mid x)$ and the upper bound $\mathcal{U}_K$ (\Cref{thm:upperboundthm}) and the lower bound $\mathcal{L}_K$ \citep{domke2018importance}, we see that by maximizing $\mathcal{L}_K$ or minimizing $\mathcal{U}_K$ we optimize different objectives w.r.t. $\tau$ (see \Cref{lem:omega_dist} for the definition of $\omega$):
\begin{align*}
\tau^* &= \argmin_{\tau \in \mathcal{F}} \mathcal{U}_{K}
= \argmin_{\tau \in \mathcal{F}} D_{KL}(q(\psi_0 \mid z) \tau(\psi_{1:K} \mid z) \mid\mid \omega_{\tau}(\psi_{0:K} \mid z))
\\
\tau_-^* &= \argmax_{\tau \in \mathcal{F}} \mathcal{L}_{K}
= \argmin_{\tau \in \mathcal{F}} D_{KL}(\omega_{\tau}(\psi_{0:K} \mid z) \mid\mid q(\psi_0 \mid z) \tau(\psi_{1:K} \mid z))
\end{align*}
In case $K = 0$ we have $\omega_{\tau}(\psi_{0:K} \mid x, z) = \tau(\psi_{0} \mid x, z)$, and the gaps become simple forward and reverse KL divergences between $\tau(\psi \mid x, z)$ and the true inverse model $q(\psi \mid z, x)$. Thus unless $\tau$ (or $\rho$) is able to represent the true inverse model exactly, one should avoid using the same distribution in both marginal log-density bounds and KL bounds \eqref{eq:kl-upper-bound} and \eqref{eq:kl-lower-bound}.

\section{Signal-to-Noise Ratio Study}\label{sec:snr-details}
In this section we provide additional details to \cref{sec:snr-intro}.

\subsection{Doubly Reparametrized Gradient Derivation}
Consider a VAE setup with multisample bound \eqref{eq:iw-bound}. Learning in such a model is equivalent to maximizing the following objective w.r.t. generative model's parameters $\theta$, inference network's parameters $\phi$, and auxiliary inference network $\tau$'s parameters $\eta$:

\begin{align*}
\mathcal{L}(\theta, \phi, \eta)
&=
\E_{\substack{q_\phi(\psi_{1:M,0}, z_{1:M} \mid x) \\ \tau_\eta(\psi_{1:M,1:K} \mid x, z_{1:M})}}
\log \frac{1}{M} \sum_{m=1}^M \frac{p_\theta(x, z_m)}{\frac{1}{K+1} \sum_{k=0}^K \frac{q_\phi(z_m, \psi_{m,k} \mid x)}{\tau_\eta(\psi_{m,k} \mid x, z_m)} }
\\
&=
\E_{\substack{q_\phi(\psi_{1:M,0}, z_{1:M} \mid x) \\ \tau_\eta(\psi_{1:M,1:K} \mid x, z_{1:M})}}
\LSE_{m=1}^M
\overbrace{
\left[
\log p_\theta(x \mid z_m) - \LSE_{k=0}^K \underbrace{\left( \log \tfrac{q_\phi(z_m, \psi_{m,k} \mid x)}{p_\theta(z_m) \tau_\eta(\psi_{m,k} \mid x, z_m)}\right)}_{\beta_{mk}}
\right]
}^{\alpha_m} + \text{const}
\end{align*}

Where $\LSE$ is a shorthand for the log-sum-exp operator, and the omitted constant is $\log \tfrac{K+1}{M}$. We will now consider a \textit{reparametrized} gradient w.r.t. $\tau$'s parameters $\nabla_\eta \mathcal{L}$, notice the REINFORCE-like term still sitting inside, and carve it out by applying the reparametrization the second time in the same way as in \citep{tucker2018doubly}. In the derivation we'll use $\softmax$ notation to mean the softmax $\tfrac{1}{1+\exp(-x)}$ function, and $\nabla_\eta \psi$ is a shorthand for $(\nabla_\eta g(\varepsilon, \eta))|_{\varepsilon = g^{-1}(\psi, \eta)}$ where $g$ is $\tau_\eta(\psi \mid x, z)$'s reparametrization.

\begin{align*}
\nabla_\eta
\mathcal{L}(\theta, \phi, \eta)
& =
\E_{q_\phi(\psi_{1:M,0}, z_{1:M})}
\nabla_\eta
\E_{\tau_\eta(\psi_{1:M,1:K} \mid z_{1:M})}
\mathop{\text{L}{\Sigma}\text{E}}_{m=1}^M[\alpha_m] \\
& =
\E_{q_\phi(\psi_{1:M,0}, z_{1:M})}
\Biggl(
\E_{\tau_\eta(\psi_{1:M,1:K} \mid z_{1:M})}
\nabla_\eta \mathop{\text{L}{\Sigma}\text{E}}_{m=1}^M[\alpha_m]
\\ & \quad\quad\quad\quad\quad\quad\quad\quad\quad
+
\E_{\tau_\eta(\psi_{1:M,1:K} \mid z_{1:M})}
\sum_{m=1}^{M} \sum_{k=1}^K
\nabla_{\psi_{mk}} \left( \mathop{\text{L}{\Sigma}\text{E}}_{m=1}^M[\alpha_m] \right) \nabla_{\eta} \psi_{mk}
\Biggr) \\
& =
\E_{q_\phi(\psi_{1:M,0}, z_{1:M})}
\left(
\E_{\tau_\eta(\psi_{1:M,1:K} \mid z_{1:M})}
\sum_{m=1}^M \softmax(\alpha)_m \nabla_\eta \alpha_m + A
\right) \\
& =
\E_{q_\phi(\psi_{1:M,0}, z_{1:M})}
\left(
\E_{\tau_\eta(\psi_{1:M,1:K} \mid z_{1:M})}
\sum_{m=1}^M \softmax(\alpha)_m \left[ -\sum_{k=0}^K \softmax(\beta_m)_k \nabla_\eta \beta_{mk} \right]
+
A
\right) \\
& =
\E_{q_\phi(\psi_{1:M,0}, z_{1:M})}
\left(
\E_{\tau_\eta(\psi_{1:M,1:K} \mid z_{1:M})}
\sum_{m=1}^M \softmax(\alpha)_m \left[ {\scriptstyle \sum_{k=0}^K \softmax(\beta_m)_k \nabla_\eta \log \tau_\eta(\psi_{mk}) } \right]
+
A
\right) \\
& =
\E_{q_\phi(\dots)}
\left(
\E_{\tau_\eta(\dots)}
\sum_{m=1}^M \sum_{k=1}^K \overbrace{\color{red} \softmax(\alpha)_m \softmax(\beta_m)_k \nabla_\eta \log \tau_\eta(\psi_{mk})}^\text{REINFORCE-like term}
+
B
+
A
\right) \\
\end{align*}
Where
\begin{align*}
A
&=
-
\E_{\tau_\eta(\psi_{1:M,1:K} \mid z_{1:M})}
\sum_{m=1}^{M} \sum_{k=1}^K
\softmax(\alpha)_m \softmax(\beta_m)_k \nabla_{\psi_{mk}} \beta_{mk} \nabla_{\eta} \psi_{mk}
\\
B
&=
\E_{\tau_\eta(\psi_{1:M,1:K} \mid z_{1:M})}
\sum_{m=1}^M \softmax(\alpha)_m \softmax(\beta_m)_0 \nabla_\eta \log \tau_\eta(\psi_{m0})
\\
q_\phi(\dots) &= q_\phi(\psi_{1:M,0}, z_{1:M}),
\quad\quad\quad
\tau_\eta(\dots) = \tau_\eta(\psi_{1:M,1:K} \mid z_{1:M})
\end{align*}

One can see the red term is the REINFORCE derivative of the $\E_{\tau_\gamma(\psi_{m,k})} \softmax(\alpha)_m \softmax(\beta_m)_k$ w.r.t. $\gamma$ with all other $\psi$ being fixed. We then evaluate this REINFORCE gradient at $\gamma = \eta$, this substitution "trick" is needed to avoid differentiating $\alpha$ and $\beta$ w.r.t. $\eta$, only their gradients w.r.t. $\psi_{mk}$ matter. We would also like to notice that though $B$ also contains $\nabla_\eta \log \tau_\eta(\psi_{m0})$ term similar to REINFORCE, it's not a REINFORCE gradient, as $\psi_{m0}$ comes from a different distribution, and thus we can not apply reparametrization to it.

\begin{align} \label{eq:dreg-tau-grad}
\nabla_\eta
\mathcal{L}(\theta, \phi, \eta)
& =
\E_{q_\phi(\dots)}
\left(
\sum_{m=1}^M \sum_{k=1}^K
\E_{\tau_\eta(\psi_{1:M,1:K \backslash mk} \mid z_{1:M})}
\left(
{
\scriptstyle
\nabla_\gamma
\E_{\tau_\gamma(\psi_{mk} \mid z_{m})}
\softmax(\alpha)_m \softmax(\beta_m)_k
}
\right)\Bigl|_{\gamma = \eta}
+
B
+
A
\right) \nonumber \\
& =
\E_{q_\phi(\dots)}
\left(
\sum_{m=1}^M \sum_{k=1}^K
{
\scriptstyle
\E_{\tau_\eta(\psi_{1:M,1:K} \mid z_{1:M})}
\nabla_{\beta_{mk}}
\left[ \softmax(\alpha)_m \softmax(\beta_m)_k \right]
 \nabla_{\psi_{mk}} \beta_{mk}
\nabla_{\eta} \psi_{mk}
}
+
B
+
A
\right) \nonumber \\
& =
\E_{\substack{q_\phi(\dots) \\ \tau_\eta(\dots)}}
\left(
\sum_{m=1}^M \sum_{k=1}^K
{
\scriptstyle
\softmax(\alpha)_m \softmax(\beta_m)_k
\left[
    1 - \softmax(\beta_m)_k (2 - \softmax(\alpha)_m)
\right]
\nabla_{\psi_{mk}} \beta_{mk}
\nabla_{\eta} \psi_{mk}
}
+
B
+
A
\right) \nonumber \\
& =
\E_{\substack{q_\phi(\dots) \\ \tau_\eta(\dots)}}
\left(
\sum_{m=1}^M \sum_{k=1}^K
\softmax(\alpha)_m
\softmax(\beta_m)_k^2
(\softmax(\alpha)_m - 2)
\nabla_{\psi_{mk}} \beta_{mk}
\nabla_{\eta} \psi_{mk}
+
B
\right) \nonumber \\
& =
\E_{\substack{q_\phi(\dots) \\ \tau_\eta(\dots)}}
\sum_{m=1}^M
\softmax(\alpha)_m
\left(
{
\scriptstyle
(\softmax(\alpha)_m - 2)
\sum\limits_{k=1}^K
\softmax(\beta_m)_k^2
\nabla_{\psi_{mk}} \beta_{mk}
\nabla_{\eta} \psi_{mk}
+
\softmax(\beta_m)_0
\nabla_\eta \log \tau_\eta(\psi_{m0})
}
\right)
\end{align}

We call this gradient estimator IWHVI-DReG. Similar derivations can be made w.r.t. $q$'s parameters $\phi$ to combat decreasing signal-to-noise ratios identified by \citet{Rainforth2018TighterVB}. We will not provide them here, as this is outside of the scope of the present work. It's also straightforward to derive a similar gradient estimator for $\rho$'s parameters in case of hierarchical prior $p(z)$, since this case essentially corresponds to nested IWAE.

\subsection{Experiments}
\begin{figure}
  \centering
  \begin{subfigure}[b]{0.65\textwidth}
    \includegraphics[width=\textwidth]{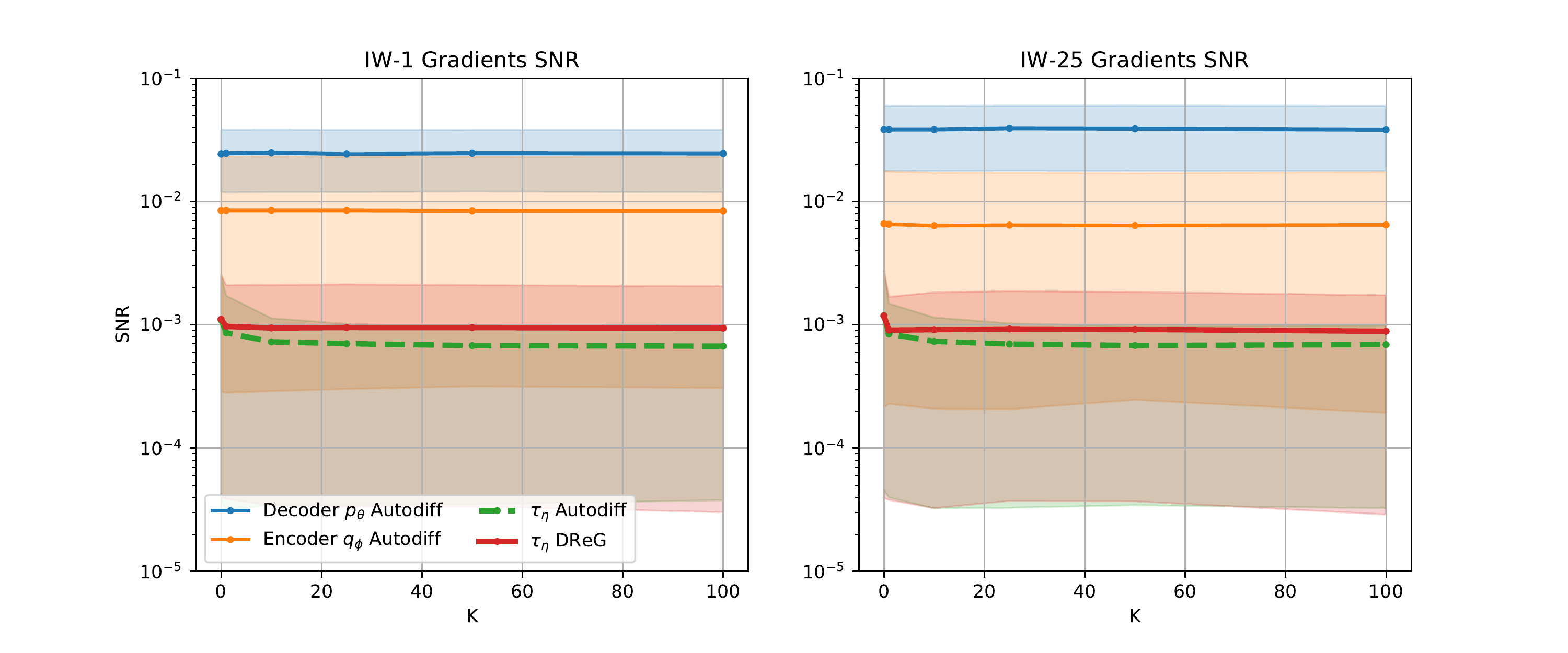}
    \caption{IWHVI-VAE for $M=1$ and $M=25$}
    \label{fig:snr-exp-vae}
  \end{subfigure}
  \begin{subfigure}[b]{0.325 \textwidth}
    \includegraphics[width=\textwidth]{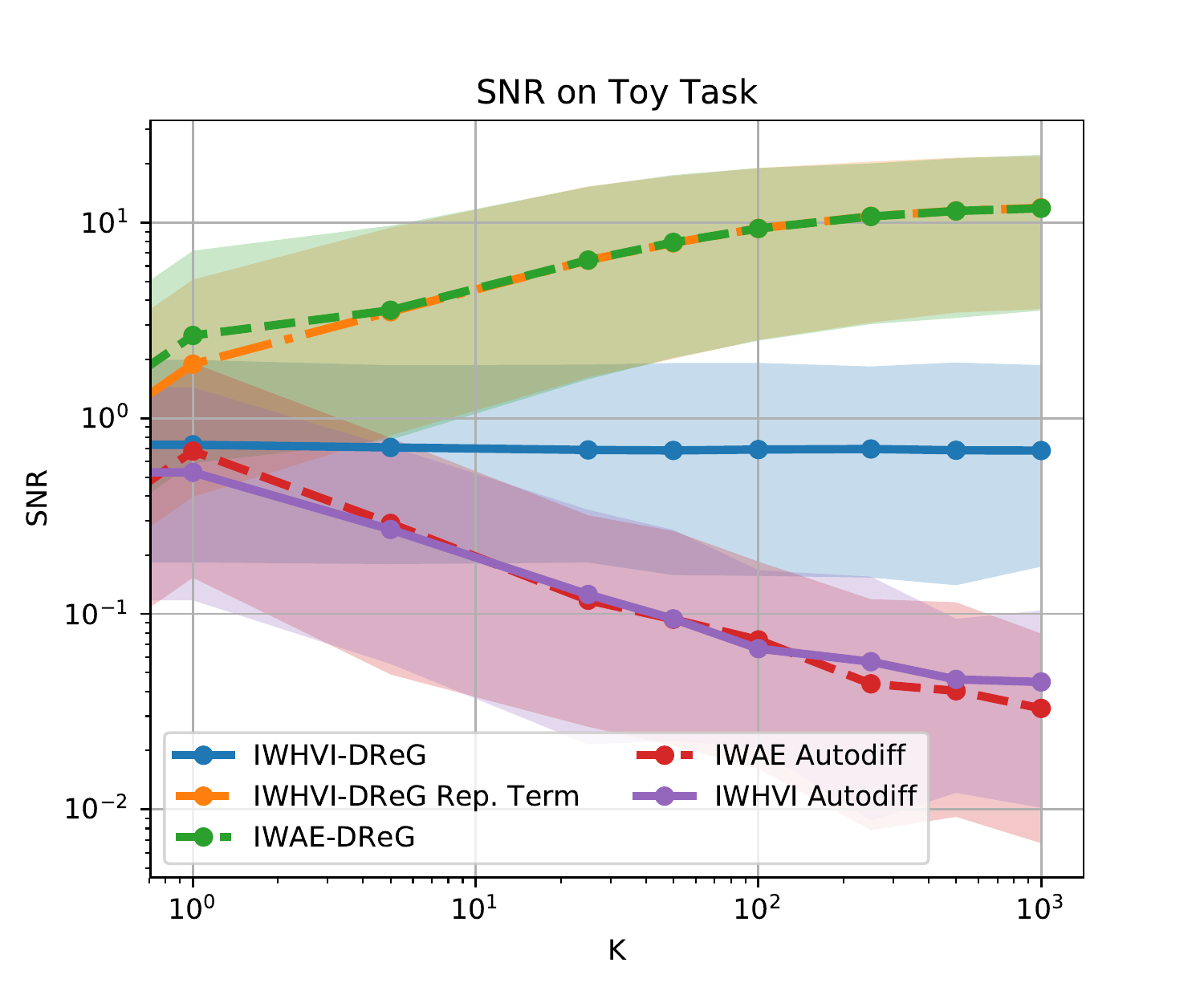}
    \caption{Toy Task}
    \label{fig:snr-exp-toy}
  \end{subfigure}
  \caption{Signal-to-Noise Ratio of gradients for different values of $K$. Solid lines denote SNR averaged over all model's parameters, and shaded area marks 90\% confidence interval over all parameters.}
\end{figure}

To evaluate the \eqref{eq:dreg-tau-grad} gradient estimate we take a slightly trained (for 50 epochs by the usual gradient) VAE and compare signal-to-noise ratios of different gradients while varying $K$ and $M$. For each minibatch we recompute the vanilla autodiff gradients and a doubly reparametrized gradient \eqref{eq:dreg-tau-grad} 100 times to estimate signal-to-noise ratio for each weight. We then average SNRs over different minibatches of fixed size, and present mean and 90\% confidence intervals over different choices of weight in \cref{fig:snr-exp-vae}.

%\begin{figure}[bt]
%   \centering
%   
%   \includegraphics[width=0.25 \textwidth]{}
%   \captionof{figure}{Signal-to-noise ratios of gradients of different model's parts for different number of IWAE samples $M$ (1, 10 and 25) and IWHVI samples $K$. Shaded area denotes 90 \% confidence interval over a choice of a single parameter.}
%   
%   \label{fig:snr-exp}
%\end{figure}

We also compute SNR on a toy task from \citep{Rainforth2018TighterVB} for an upper bound (\cref{thm:upperboundthm}) on $\E_{q(x)} \log q(x)$ for ${q(x, z) = \mathcal{N}(x \mid z, I) \mathcal{N}(z \mid \theta, I)}$ and $\tau_\eta(z \mid x) = \mathcal{N}(z \mid Ax+b, 2/3)$ for $\theta = [1, \dots, 1], x, z  \in \mathbb{R}^{10}$. We first train $\eta = \{A, b\}$ to optimality by making 1000 AMSGrad \citep{j.2018on} steps with learning rate $10^{-2}$ with batches of 100, then evaluate the gradients 1000 times with batches of size 100. We also include gradients of IWAE (coming from the task of lower bounding the $\E_{q(x)} \log q(x)$) to compare with. We compute SNR per parameter over these 1000 samples, and present results in \cref{fig:snr-exp-toy}.

Best seen in the toy task, SNR of autodiff gradients decreases as $K$ grows, much resembling the standard IWAE issue, outlined by \citep{Rainforth2018TighterVB}. IWHVI-DReG solves the problem, but only partially. While the SNR is no longer decreasing, it does not increase as in IWAE-DReG. While detailed study of this phenomena is outside the scope of this work, we show that it's the second term (denoted $B$) in IWHVI-DReG, coming from the $\psi_{m0}$-based term in the bound, that causes the trouble. Indeed, if we omit this term, the (now biased) estimate would have higher SNR, essentially approaching the standard IWAE's one. In practice, we found that the improved gradient estimate IWHVI-DReG significantly increased $D_{KL}(\tau(\psi \mid z)\mid\mid q(z))$, but this resulted in minor increases in the validation log-likelihood. 

\section{Debiasing the bound}\label{sec:jhvi-details}
\subsection{Deriving the bound}

Following \citep{nowozin2018debiasing} we argue (\cref{lem:uk_asymp_expansion}) that

$$
\mathcal{U}_{K}
=
\E_{p(\psi_0 \mid z)} \E_{\tau(\psi_{1:K} \mid z)} \log \left( \frac{1}{K+1} \sum_{k=0}^{K} \frac{p(z, \psi_k)}{\tau(\psi_k \mid z)} \right)
= \log p(z)
+
\sum_{j=1}^\infty
\frac{\gamma_j}{(K+1)^j}
$$

So $\mathcal{U}_{K}$ can be seen as a biased estimator of marginal log-density $\log p(z)$ with bias of order $O(1/K)$. We can reduce this bias further by making use of the following fact:
\begin{align*}
(K+1) \mathcal{U}_{K} - K \mathcal{U}_{K-1}
&= \log p(z) - \frac{\gamma_2}{K (K+1)} + O\left(\tfrac{1}{K^2}\right) \\
&= \log p(z) - \sum_{j=0}^\infty \frac{\gamma_2}{(K+1)^{j+2}} + O\left(\tfrac{1}{K^2}\right) \\
&= \log p(z) - \frac{\gamma_2}{(K+1)^2} + O\left(\tfrac{1}{K^2}\right)
\end{align*}
Averaging this identity over all possible choices of a subset of samples from $\tau(\psi \mid z)$ of size $K-1$, we obtain Jackknife-corrected estimator with bias of order $O(1/K^2)$. One could then apply the same procedure again and again to obtain Generalized-order-$J$-Jackknife-corrected estimator with bias of order $O(1/K^{J+1})$, leading to a $J$-Jackknife upper marginal log-density estimate:
\begin{align} \label{eq:jhvi}
\mathcal{J}_{K}^J
=
\sum_{j=0}^J c(K, J, j) \overline{\mathcal{U}}_{K-j}
\end{align}
Where $\overline{\mathcal{U}}_{K-j}$ is $(K-j)$-samples bound averaged over all possible choices of a subset of size $K-j$ from $\psi_{1:K}$, and $c(K, J, j)$ are Sharot coefficients \citep{sharot1976generalized,nowozin2018debiasing}:
$$
\overline{\mathcal{U}}_{K-j}
=
\frac{1}{ \binom{K}{K-j} }
\sum_{ S \subseteq \{1, \dots, K\}: |S| = K-j }
\log \left( \frac{1}{K-j+1} \left[ \frac{p(z, \psi_0)}{\tau(\psi_0 \mid z)} + \sum_{k \in S} \frac{p(z, \psi_k)}{\tau(\psi_k \mid z)} \right] \right)
$$

$$
c(K, J, j) = (-1)^j \frac{(K-j)^J}{(J-j)! j!}
$$

Despite not guaranteed to be an upper bound anymore, we still see that the estimate tends to overestimate the marginal log-density in practice.

\begin{lemma}\label{lem:uk_asymp_expansion}
For fixed $p(\psi_0 \mid z), \tau(\psi_{1:K} \mid z)$ and $z$ there exists a sequence $\{\gamma_k\}_{k=1}^\infty$ s.t.

\begin{align} \label{eq:uk_bias_decomp}
\mathcal{U}_{K}
= \log p(z)
+
\sum_{j=1}^\infty
\frac{\gamma_j}{(K+1)^j}
\end{align}
\begin{proof}

First, we note that $\mathcal{U}_{K}$ can be represented as marginal log-density plus some non-negative (due to \cref{thm:upperboundthm}) bias, which we'll consider in greater detail.
$$
\mathcal{U}_{K}
=
\E_{\substack{p(\psi_0 \mid z) \\ \tau(\psi_{1:K} \mid z)}} \log \left( \frac{1}{K+1} \sum_{k=0}^{K} \frac{p(z, \psi_k)}{\tau(\psi_k \mid z)} \right)
=
\log p(z)
+
\overbrace{
\E_{\substack{p(\psi_0 \mid z) \\ \tau(\psi_{1:K} \mid z)}} \log \left( \frac{1}{K+1} \sum_{k=0}^{K} \frac{p(\psi_k \mid z)}{\tau(\psi_k \mid z)} \right)
}^\text{Bias}
$$

Denote $w_k = \tfrac{p(\psi_k \mid z)}{\tau(\psi_k \mid z)}$, $w_k' = w_k - 1$ and expand the Bias around 1:

\begin{align*}
\text{Bias}
& =
\sum_{n=1}^\infty \frac{(-1)^{n + 1}}{n} \E_{p(\psi_0 \mid z)} \E_{\tau(\psi_{1:K} \mid z)} \left( \tfrac{1}{K+1} \sum_{k=0}^K w_k' \right)^n \\
&=
\sum_{n=1}^\infty \frac{(-1)^{n + 1}}{n} \E_{p(\psi_0 \mid z)} \E_{\tau(\psi_{1:K} \mid z)}  \left( \tfrac{1}{K+1} w_0' + \tfrac{K}{K+1} \overline{w_{1:K}'} \right)^n \\
& =
\sum_{n=1}^\infty \frac{(-1)^{n + 1}}{n} \sum_{m=0}^n {n \choose m} \E_{p(\psi_0 \mid z)} \left( \tfrac{w_0'}{K+1} \right)^m  \E_{\tau(\psi_{1:K} \mid z)} \left( \tfrac{K}{K+1} \overline{w_{1:K}'} \right)^{n-m} \\
& =
\sum_{n=1}^\infty \frac{(-1)^{n + 1}}{n} \sum_{m=0}^n {n \choose m}
\left(1 - \tfrac{1}{K+1}\right)^{n-m} \left(\tfrac{1}{K+1}\right)^n
\E_{p(\psi_0 \mid z)} \left( w_0' \right)^m \E_{\tau(\psi_{1:K} \mid z)} \left( \overline{w_{1:K}'} \right)^{n-m}
\end{align*}

Where $\overline{w_{1:K}'} = \tfrac{1}{K} \sum_{k=1}^K w_k'$ -- an empirical average of zero-mean random variables. \citep{angelova2012moments} has shown that for a fixed $s \in \mathcal{N}$ there exist $\gamma_1^{(s)}, \dots, \gamma_T^{(s)}$ s.t.

$$
\E_{\tau(\psi_{1:K} \mid z)} \left( \overline{w_{1:K}'} \right)^s
= \sum_{t=1}^T \frac{\gamma_t^{(s)}}{K^t}
= \sum_{t=1}^T \gamma_t^{(s)} \left[ \sum_{n=1}^\infty \frac{1}{(K+1)^n} \right]^t
$$

Therefore every addend in Bias depends on $K$ only through $1/(K+1)$, and thus decomposition \eqref{eq:uk_bias_decomp} holds.

Note on convergence: while we have not shown the presented series to be convergent, in practice we only care about bias'  asymptotic behaviour up to an order of $1/(K+1)^J$, for which the decomposition works as long as the corresponding moments in Bias exist and are finite.

\end{proof}
\end{lemma}

\subsection{Experiments}
\begin{figure}
  \centering
    \includegraphics[width=0.5 \textwidth]{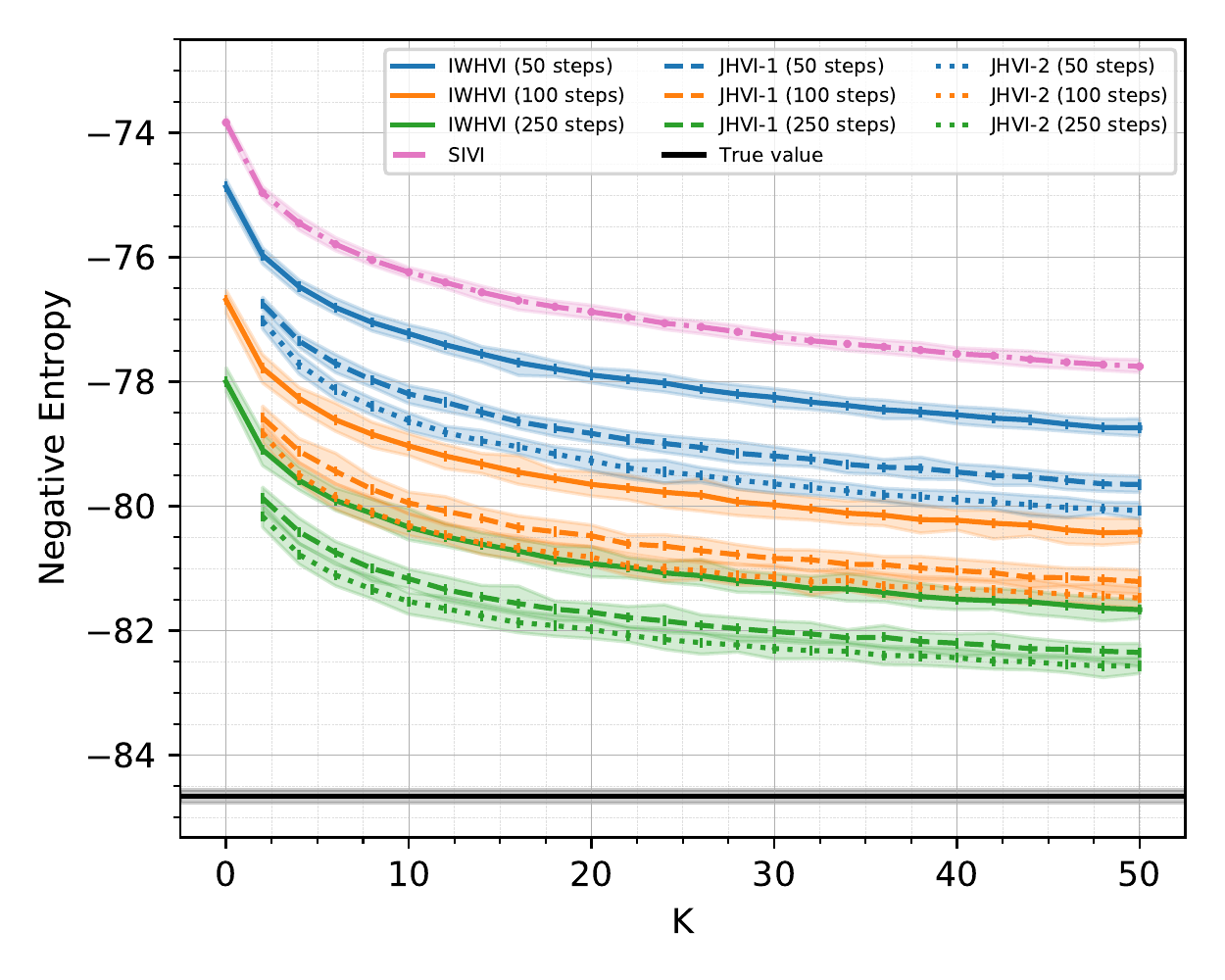}
    \label{fig:toy-exp-jk}
  \caption{Negative entropy bound for 50-dimensional Laplace distribution. Shaded area denotes 90\% confidence interval.}
\end{figure}

We only perform experimental validation of the Jackknife upper KL estimate (JHVI for short) in the same setting as in \cref{sec:toy-exp}. \cref{fig:toy-exp-jk} shows improved performance of the Jackknife-corrected estimate.

\section{Experiments Details} \label{sec:vae-exp-details}
For MNIST we follow the setup by \citet{mescheder2017adversarial}: we use single 32-dimensional stochastic layer with ${p(z)} = {\mathcal{N}(z \mid 0, I)}$ prior, decoder ${p_\theta(x \mid z)} = {\text{Bernoulli}(x \mid \pi_\theta(z))}$ where $\pi_\theta$ is a neural network with two hidden 300-neurons layers and a softplus nonlinearity \footnote{This is the nonlinearity used in \citet{mescheder2017adversarial}'s code for fully-connected experiments.}, and latent variable model encoder ${q_\phi(z \mid x)} = \int {\mathcal{N}(z  \mid \mu_\phi(x, \psi), \sigma_\phi^2(x, \psi))} {\mathcal{N}(\psi \mid 0, 1)} {d \psi}$ where ${\mu_\phi(x, \psi)}$ and ${ \sigma_\phi^2(x, \psi)}$ are outputs of a neural network with architecture similar to as that of the decoder, except each next layer (including the one that generates distribution's parameters) acts on previous layer's output concatenated with input $\psi$. We take $\tau_\vartheta(\psi \mid z, x) = \mathcal{N}(\psi  \mid \nu_\vartheta(x, z), \varsigma_\vartheta^2(x, z))$ where mean and variance are outputs of another neural network with same network architecture as that of the decoder, except it takes concatenation of $x$ and $z$ as input.

For OMNIGLOT we used simiar architecture, but for 50-dimensional $z$ and $\psi$, and all hidden layers had 200 neurons.

For flow-based models we used their implementations from TensorFlow Probability \citep{dillon2017tensorflow}. We had the encoder network output not only $\mu$ and $\sigma$, but also a context vector $h$ of same size as $z$ to be used to condition the flow transformations.

For IWHVI we used a grid search over 1) whether to use IWHVI-DReG \eqref{eq:dreg-tau-grad}, 2) whether to do inner KL (the $\log \tfrac{q(\psi)}{\tau(\psi \mid x, z)}$ terms) warm-up (linearly increase its weight from 0 to 1) for first 300 epochs, 3) whether to do outer KL warmup (the $\log \tfrac{p(z)}{\tfrac{1}{K+1} \sum \dots}$ term) for first 300 epochs, 4) Either use learning rate $\eta_\text{base} = 10^{-3}$ with annealing $\eta = 0.95^{\text{epoch}/100} \eta_\text{base}$ or $\eta = 10^{-4}$ without annealing. We used the same grid for SIVI, except options (1) and (2) had no effect, since they only influenced $\tau$, and thus we did not serach over them. For HVM only the option (1) had no effect, since HVM does not sample from $\tau$.

In all experiments we did $10,000$ epochs with batches of size 256 and used Adam \citep{DBLP:journals/corr/KingmaB14} optimizer with $\beta_1=0.9, \beta_2=0.999$. We held out $5,000$ examples from trainsets to be used as validation.

\end{document}